%% file: acl_latex.tex
\pdfoutput=1

\documentclass[11pt]{article}

\usepackage[final]{acl}

\usepackage{times}
\usepackage{latexsym}

\usepackage[T1]{fontenc}

\usepackage[utf8]{inputenc}

\usepackage{microtype}

\usepackage{inconsolata}

\usepackage{graphicx}
\usepackage{multirow}
\usepackage{multicol}
\usepackage{booktabs}
\usepackage{tipa}
\usepackage{amsmath}
\usepackage{amssymb}
\usepackage{colortbl}
\usepackage{tipa}
\usepackage{graphicx}
\usepackage{subcaption}
\usepackage{enumitem}
\usepackage{xcolor}

\definecolor{LightBlue}{rgb}{0.8235,0.9737,0.9882}
\definecolor{YellowGreen}{rgb}{0.6039, 0.8039, 0.1961}
\usepackage{tablefootnote,footnotehyper}

\title{Cross-Lingual IPA Contrastive Learning for Zero-Shot NER}

\author{Jimin Sohn \\
  LG Innotek, South Korea \\
  \texttt{jimin.sohn@lginnotek.com} \\\And
  David R. Mortensen \\
  Carnegie Mellon University, USA \\
  \texttt{dmortens@cs.cmu.edu} \\}

\begin{document}
\maketitle
\begin{abstract}
Existing approaches to zero-shot Named Entity Recognition (NER) for low-resource languages have primarily relied on machine translation, whereas more recent methods have shifted focus to phonemic representation. Building upon this, we investigate how reducing the phonemic representation gap in IPA transcription between languages with similar phonetic characteristics enables models trained on high-resource languages to perform effectively on low-resource languages.
In this work, we propose \textbf{CON}trastive \textbf{L}earning with \textbf{IPA} (\textbf{CONLIPA}) dataset containing 10 English and high resource languages IPA pairs from 10 frequently used language families. We also propose a cross-lingual \textbf{IPA} \textbf{C}ontrastive learning method (\textbf{IPAC}) using the CONLIPA dataset. Furthermore, our proposed dataset and methodology demonstrate a substantial average gain when compared to the best performing baseline.
\end{abstract}

\section{Introduction}

One of the facts that links the languages of the world together is shared vocabulary. Languages that are phylogenetically related to one another inherit shared words (cognates) and languages that are in contact with one another borrow words (loanwords) from one another. These etymologically related words tend to share similar meanings and similar pronunciations. Various attempts have been made to leverage this similarity. For example, \citet{bharadwaj-etal-2016-phonologically} used phonetic feature representations of Uyghur and Turkish to leverage shared names in Named Entity Recognition (NER) and \citet{chaudhary-etal-2018-adapting} used IPA (International Phonetic Alphabet) representation to improve NER and machine translation in Bengali (pivoting from Hindi). However, these past approaches have proposed models that learned representations for phoneme strings. Loanwords or cognates have similar embedded representations because their IPA representations are similar. We propose, instead, to learn representations---using contrastive learning---that capture the phonological aspects of etymologically-related words across languages.

\begin{figure*}[t!]
    \centering
    \includegraphics[width=1.0\textwidth]{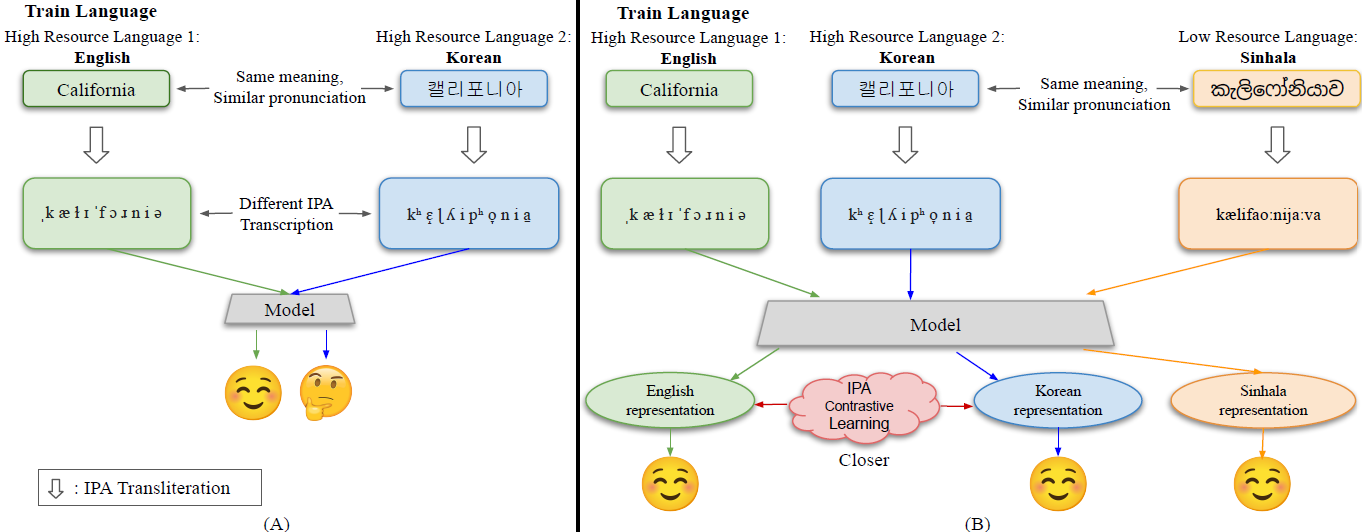}
    \caption{Concept Figure. As shown in (A), existing phonemic models struggle to recognize the same word when IPA representations differ across languages, despite similar pronunciations. In contrast, our method (B) uses IPA contrastive learning to align representations of languages with similar pronunciations, particularly for high-resource languages. This enables effective zero-shot inference for low-resource languages, demonstrating strong generalization.}
    \label{fig:concept_fig}
\end{figure*}

Various approaches for zero-shot NER in low-resource languages, where data acquisition is challenging, have been proposed over time.
Most previous methods~\cite{yang-etal-2022-crop,liu-etal-2021-mulda,mo2024mcl} often employed machine translation with grapheme-based inputs.
Since machine translation utilizes prior knowledge of low-resource languages, a method using phonemic representation was proposed for a stricter zero-shot setting \cite{sohn-etal-2024-zero}.

We investigate how reducing the phonemic representation gap in IPA transcription between languages with similar phonetic characteristics enables models trained on high-resource languages to perform effectively on low-resource languages as shown in Figure \ref{fig:concept_fig}. We selected 10 representative languages from 10 widely spoken language families and collected IPA pairs with English that share the same meaning and similar pronunciation. Using this \textbf{CON}trastive \textbf{L}earning with \textbf{IPA} (\textbf{CONLIPA}) dataset, we conducted Cross-lingual \textbf{IPA} \textbf{C}ontrastive learning method (\textbf{IPAC}) on the phonemic representation space. Extensive experiments and cosine similarity score demonstrate that our method effectively brings the representations of similarly pronounced words across different languages closer together.

Our approach differs from \cite{sohn-etal-2024-zero} in that the model is explicitly trained to represent IPA in a cross-linguistically meaningful way. It is not merely about token overlap; the model learns to represent phonetically transcribed words in a manner that ensures similarity with etymologically related words, such as named entities, in other languages. \cite{zouhar-etal-2024-pwesuite} also employed similar techniques, including metric learning and triplet margin loss, to learn neural representations of IPA strings. However, their approach was monolingual in nature, as both positive and negative samples were drawn from the same language as the anchor, and the metric space was defined based on phonetic features.

We also explored the interesting feature of the Korean language, which allows foreign pronunciations to be recorded using \emph{Hangul} in a way that closely approximates the original pronunciation. Leveraging this feature, we highlight the potential of Korean for future zero-shot NER research.

In general, the main contributions of this paper are as follows:
\begin{itemize}[nosep]
\item We propose the \textbf{CON}trastive \textbf{L}earning with \textbf{IPA} (\textbf{CONLIPA}) dataset, which contains IPA pairs of English and 10 languages from 10 widely spoken language families.
\item We propose a novel Cross-Lingual \textbf{IPA} \textbf{C}ontrastive Learning (\textbf{IPAC}) approach using the CONLIPA dataset, aimed at reducing the gap in phonemic representations between high-resource languages with similar pronunciations.
\item We investigate \textbf{Unimodal Contrastive Learning} using exclusively phonemic input, without incorporating multimodal inputs such as images or audio.
\item To the best of our knowledge, we are the first to use LLMs, such as ChatGPT, to extract cognate pairs and train a model using these pairs.
\item We evaluate the proposed method using WikiANN NER dataset and compare it with baseline methods. Experimental results verify the effectiveness of our method and demonstrate its significant advantages in Zero-Shot NER with low resource language task.
\end{itemize}

\section{Related Work}
\subsection{Zero-shot Cross-lingual NER}
Zero-shot cross-lingual NER is crucial for low-resource languages, where labeled data is scarce. While previous works~\cite{yang-etal-2022-crop,liu-etal-2021-mulda,mo2024mcl} used parallel data from machine translation, this approach faces challenges for languages where machine translation is not feasible. ZGUL \cite{rathore-etal-2023-zgul} established a strict zero-shot setting with no target language data, relying on a language adapter trained on typologically similar languages. However, it uses grapheme-based input, limiting its applicability to languages with novel orthographic systems, and is restricted to specific language groups—Germanic, Slavic, African, and Indo-Aryan. In contrast, our approach covers 10 widely spoken language families and does not require overlap between the training and inference languages.

Some works~\cite{bharadwaj-etal-2016-phonologically, chaudhary-etal-2018-adapting} have utilized phonemic representation for NER, but they did not operate in a zero-shot setting. In contrast, \cite{sohn-etal-2024-zero} performed NER by using IPA phonemes as input in a strict zero-shot setting, where no data or prior knowledge was available for the inference language. However, it trained the model exclusively on English data and did not fully address discrepancies in IPA notation for languages with similar pronunciations. The XPhoneBERT\cite{thenguyen23_interspeech} backbone model used by \cite{sohn-etal-2024-zero} learns to represent phoneme strings such that similar strings have similar representations. In contrast, our CONLIPA learns to represent phoneme strings, such as names, in a way that ensures they have similar representations to phonologically and semantically related strings in other languages.

\begin{figure*}[t!]
    \centering
    \includegraphics[width=0.8\textwidth]{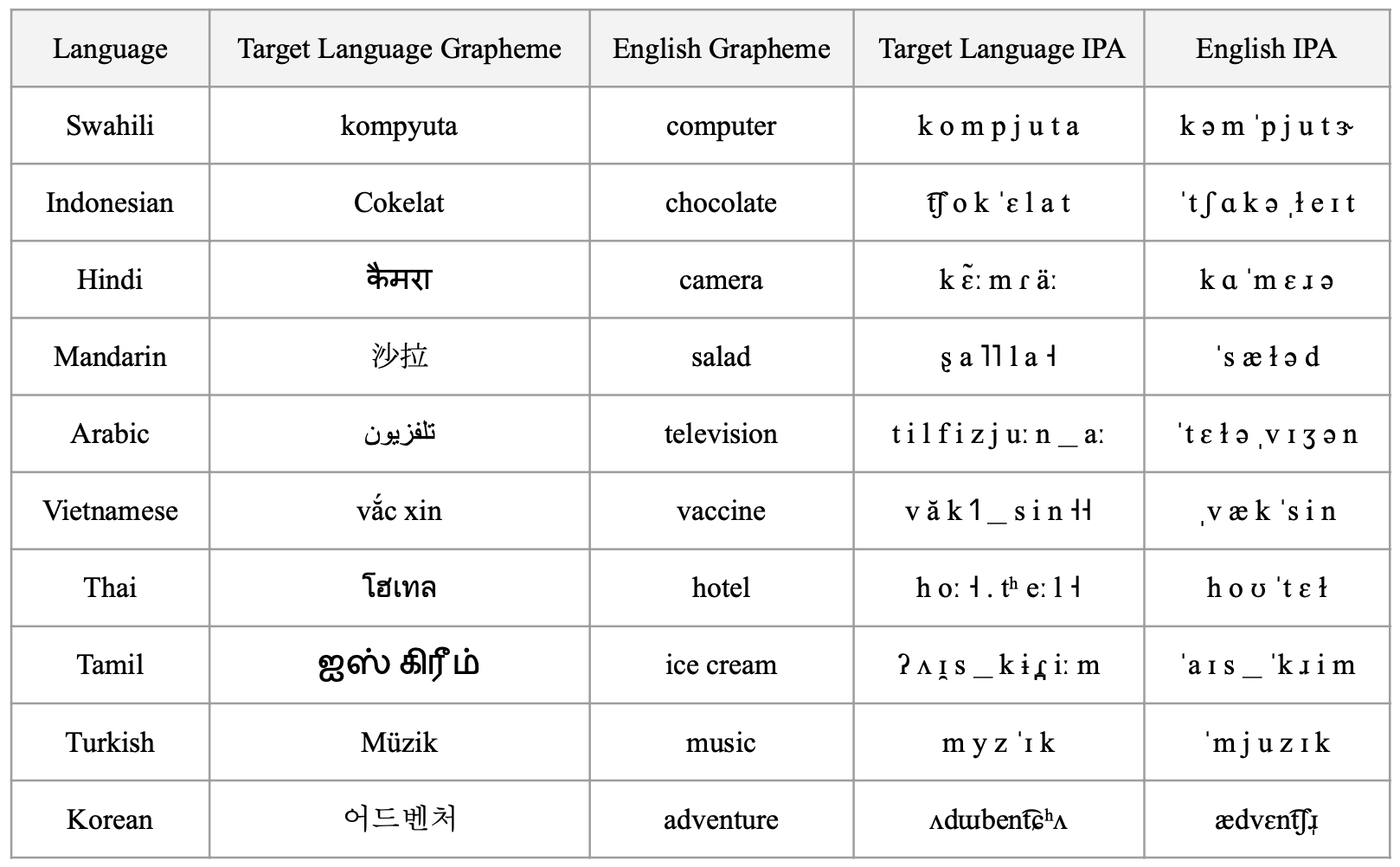}
    \caption{Samples in our CONLIPA dataset for each language.}
    \label{fig:data_sample}
\end{figure*}

\subsection{Multimodal and Unimodal Contrastive Learning}
Contrastive learning is a self-supervised approach that brings similar data points closer in feature space while pushing dissimilar points apart, enabling the learning of meaningful representations without labeled data by typically using InfoNCE loss~\cite{DBLP:journals/corr/abs-1807-03748, chen2020simple}. CLIP~\cite{radford2021learning} extends this to learn joint representations of images and text by aligning their features in a shared embedding space, advancing contrastive learning in the multimodal domain, primarily focusing on bridging image-text gaps.

As mentioned in \cite{huang2024on}, unimodal contrastive learning has generally not achieved the same level of success as the unprecedented success of multimodal contrastive learning. The foundational work on contrastive learning has explored key aspects such as alignment and uniformity of contrastive loss \cite{wang2020understanding}, the impact of auxiliary tasks on learning representations \cite{lee2021predicting}, and optimization perspectives on self-supervised learning \cite{tian2020understanding}. Additionally, several studies have analyzed contrastive learning in single-modal and multi-view settings~\cite{8507bd7339ff4b2a8b8854368006864c, haochen2021provable, tosh2021contrastive, saunshi2022understanding}. \cite{wen2021toward} study ReLU networks but differs by requiring an adjustable bias term and not considering multimodal contrastive learning. \cite{zouhar-etal-2024-pwesuite} also employed related approaches, such as metric learning and triplet margin loss, to learn neural representations of IPA strings. However, their approach was purely monolingual, with positive and negative samples drawn from the same language as the anchor and the metric space defined by phonetic features. Unlike these studies, our approach differs in that it employs a unimodal contrastive learning methodology using only phonemic input based on phonetic features across different languages in a multilingual setting.

\subsection{Contrastive Learning with Phoneme Embedding}
There have been some research on contrastive learning utilizing phoneme embedding. IPA-CLIP \cite{matsuhira2023ipa} is a multimodal method that uses image, text, and IPA, with only using English in both text and IPA. As zero-shot inference experiments on various languages were not conducted, it is difficult to guarantee strong performance across all languages, as IPA symbols may differ between English and other languages.

PLCL \cite{kewei2024phoneme} is also a multimodal approach that performs contrastive learning between English audio-audio and audio-text pairs. We note that our Cross-Lingual IPA Contrastive Learning (IPAC) clearly differentiates itself by focusing on contrastive learning between phoneme embedding of different languages, rather than within the multimodal domain.

\section{CONLIPA Dataset}
In this section we provide an overview of how we created the \textbf{CON}trastive \textbf{L}earning with \textbf{IPA} (\textbf{CONLIPA}) dataset. The dataset is used in the cross-lingual IPA contrastive learning experiments presented in Section \ref{section_4}.

\begin{table}[h!]
\centering
\resizebox{\columnwidth}{!}{
\begin{tabular}{llr}
\toprule
Language Family&Language&Data \\
\midrule
Atlantic-Congo & Swahili & 27 \\
Austronesian & Indonesian & 86 \\
Indo-European & Hindi & 128 \\
Sino-Tibetan & Mandarin & 6 \\
Afro-Asiatic & Arabic & 34 \\
Austroasiatic & Vietnamese & 10 \\
Tai-Kadai & Thai & 31 \\
Dravidian & Tamil & 71 \\
Turkic & Turkish & 52 \\
Koreanic & Korean & 7521 \\
\bottomrule
\end{tabular}
}
\caption{\label{tab:language_family}
Selected 10 language families, one of their representative Languages, and the number of data samples per each language.
}
\end{table}

\subsection{Language Selection}
We selected 10 major language families and chose one representative language from each family. These languages are high-resource, which makes it easier to obtain IPA pairs with similar phonetic characteristics between the target language and English. We selected the top 9 most widely used language families in the world (\emph{Atlantic-Congo}, \emph{Austronesian}, \emph{Indo-European}, \emph{Sino-Tibetan}, \emph{Afro-Asiatic}, \emph{Austroasiatic}, \emph{Tai-Kadai}, \emph{Drividian}, \emph{Turkic}), and added Korean from the \emph{Koreanic} language family. We included Korean because it is a well-resourced language with a strongly phonemic orthography that, like IPA, has the potential to represent other languages phonemically. This characteristic enabled us to collect a significantly larger amount of data compared to other 9 languages.

Additionally, our CONLIPA dataset used for training contains a minimum of 6 and up to 512 instances per language, enabling efficient and fast fine-tuning. Due to the relatively low computational and memory requirements, the training process incurs minimal computational cost and power consumption, making it more environmentally sustainable. The 10 selected language families, along with the representative languages from each family and the corresponding number of data samples, are presented in Table \ref{tab:language_family}.

\subsection{Dataset Creation}
We collected pairs of foreign loanwords from English and 10 representative languages that have similar meanings and pronunciations using ChatGPT\footnote{https://chatgpt.com/}. Since these languages borrow and use English words directly, the words are transcribed in the closest possible form to original English pronunciation. These words are all loanwords, so it seems that ChatGPT recognizes them as part of a translation task. Additionally, these 10 languages are high-resource languages, meaning that ChatGPT has likely been trained on a large amount of translation data for them. The words obtained were then manually verified by the authors to ensure their pronunciation similarity for each representative language, using Google Translate\footnote{https://translate.google.co.kr/} and online dictionaries. The choice of English as a reference language was motivated by its status as a high-resource language with extensive datasets in NLP, making it likely that models already possess strong representations for English.

\begin{figure*}[t!]
    \centering
    \includegraphics[width=1.0\textwidth]{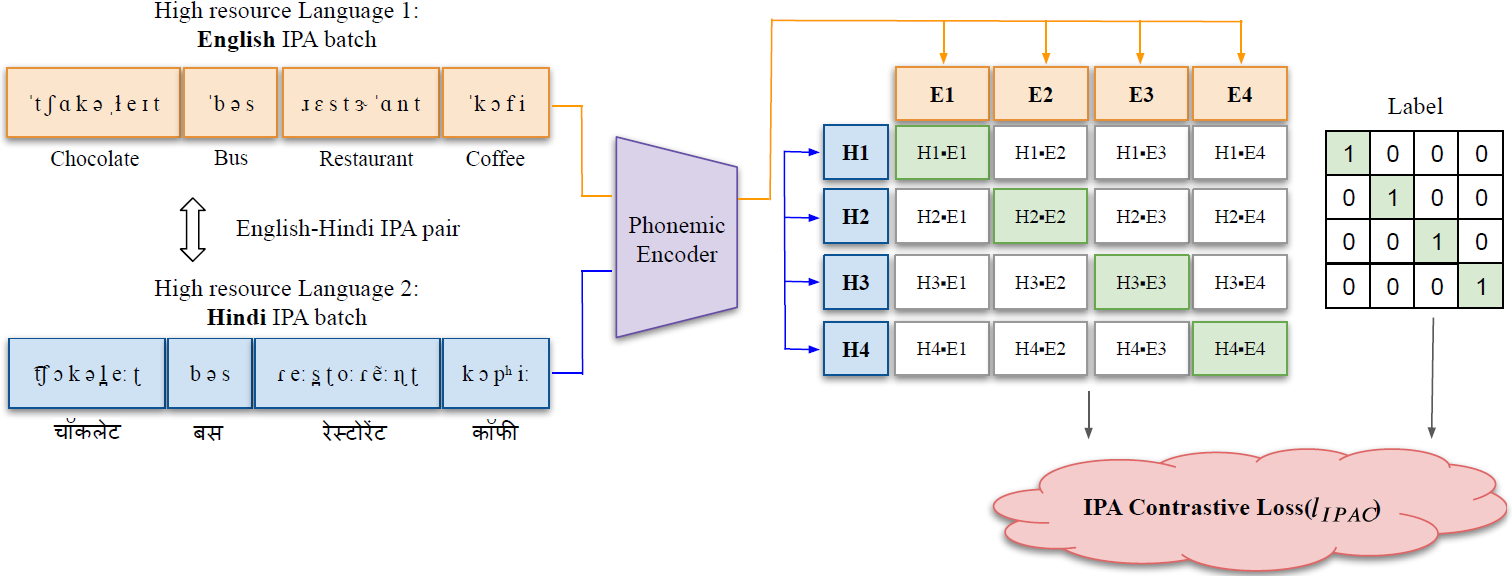}
    \caption{Overall architecture of our IPA Contrastive Learning (IPAC). First, the IPA representations of word pairs with similar pronunciations are obtained from the phonemic encoder for two high-resource languages, such as English and Hindi. Then, these pairs are considered \colorbox{YellowGreen}{positive pairs}, while the remaining samples in the batch are treated as \fcolorbox{gray}{white}{negative pairs} to compute the contrastive loss.}
    \label{fig:main_figure}
\end{figure*}

As shown in Table \ref{tab:language_family}, the number of such pairs varied significantly across languages. For instance, Mandarin had only 6 pairs due to the limited number of similar pronunciations with English, while Korean, with its ability to represent foreign words phonetically using \emph{Hangul}, allowed for a much larger collection of 7,521 samples. Through experimental evaluation, we found that using \textbf{only 512 samples} out of the 7,521 Korean samples yielded the best performance, as shown in Section \ref{sec:korean_data_num_ablation} Table \ref{tab:korean_num_case1}.

We converted the grapheme notation $G$ of English $e$ and the 10 target languages $t \in
\{ \texttt{swa}, \texttt{ind}, \texttt{hin}, \texttt{cmn}, \texttt{ara}, \texttt{vie}, \texttt{tha}, \texttt{tam}, \texttt{tur}, \texttt{kor} \}$, into IPA notation $I$. We used CharsiuG2P toolkit \cite{Zhu2022ByT5MF} which XPhoneBERT\cite{thenguyen23_interspeech} originally employed for IPA transliteration. As shown in Figure \ref{fig:data_sample}, the dataset format consist of 4 components, which are $(G_t, G_e, I_t, I_e)$.

\section{Cross-Lingual IPA Contrastive Learning (IPAC)}
\label{section_4}
Contrastive learning is a widely used self-supervised learning approach, particularly in image-text representation tasks. Its core concept involves training a model to determine whether two input samples are similar or different by evaluating them within a learned latent space.

Our approach differs in that, instead of using image-text pairs of two different modalities, we input IPA transcriptions of two different languages into a phonemic encoder. The goal is to cross-lingually align their phonemic representations. As shown in Figure \ref{fig:main_figure}, we performed IPAC by treating pairs of similar-sounding English IPA and target language IPA as positive samples from the CONLIPA dataset, while considering other samples within the batch as negative samples.

We utilized the InfoNCE loss \cite{DBLP:journals/corr/abs-1807-03748, chen2020simple} in our IPA contrastive learning framework, as it is a widely adopted loss function in contrastive learning. This loss function enhances the mutual information between positive pairs while reducing it between positive and negative pairs. The loss is defined as follows:
\begin{align} \label{eqn:InfoNCE_loss}
  \begin{split}
	l(I_e,I_t)=\frac{1}{N}\sum_{i}-log\frac{\exp{((I_e^i)^TI_t^i/\tau)}}{\sum_{k}\exp{((I_e^i)^TI_t^i/\tau)}}
  \end{split}
\end{align}
where $\tau$ is a hyperparameter called temperature coefficient, $N$ refers to the batch size, $T$ refers to the transpose of a matrix and $i$ refers to the $i^{th}$ sample of the batch.

Following the convention, we also perform $l(I_t,I_e)$ symmetrically and calculated the average as shown below, which is used as the final \textbf{IPA} \textbf{C}ontrastive loss $l_{IPAC}$.
\begin{align} \label{eqn:my_equation}
  \begin{split}
	l_{IPAC}=\frac{1}{2}(l(I_e, I_t)+l(I_t, I_e))
  \end{split}
\end{align}

\section{Experiment Setting}

\begin{table*}[ht!]
\centering
\resizebox{\textwidth}{!}{
\begin{tabular}{c|ccccccccccccccccc}
\toprule
Case & Input & Model & \multicolumn{13}{c}{Languages} & AVG & STD \\
\midrule
\multirow{4}{*}{CASE 1}&&&sin & som & mri & quy & uig & aii & kin & ilo & & & & & & & \\ \cmidrule{2-18}
 & grapheme & mBERT & 10.71 & \textbf{44.76} & 38.48 & 55.07 & 18.70 & 12.58 & \textbf{62.37} & 79.51 & & & & & & 40.27 & 25.00 \\
& grapheme & CANINE & 26.31 & 43.35 & \textbf{51.30} & \textbf{59.48} & 27.19 & 22.38 & 54.74 & \textbf{80.70} & & & & & & 45.68 & 19.99  \\
&phoneme & XPhoneBERT(baseline) & 43.61 & 38.91 & 38.07 & 51.90 & 44.82 & 31.03 & 49.67 & 73.05 & & & & & & 46.38 & 12.67  \\
&phoneme & CONLIPA(ours) & \cellcolor{LightBlue}\textbf{45.69} & 38.7 & \cellcolor{LightBlue}39.67 & \cellcolor{LightBlue}57.7 & \cellcolor{LightBlue}\textbf{45.17} & \cellcolor{LightBlue}\textbf{34.92} & \cellcolor{LightBlue}50.58 & \cellcolor{LightBlue}73.35 & & & & & & \cellcolor{LightBlue}\textbf{48.22} & \cellcolor{LightBlue}\textbf{12.44}  \\
\midrule
\multirow{4}{*}{CASE 2} & & & epo & khm & tuk & amh & mlt & ori & san & ina & grn & bel & kur & snd & &  &  \\
\cmidrule{2-18}
& grapheme & mBERT & 71.31 & 16.12 & \textbf{64.52} & 11.90 & \textbf{63.83} & 9.96 & 48.73 & \textbf{73.89} & 50.44 & \textbf{83.12} & 54.16 & 35.02 & & 48.58 & 25.13 \\
& grapheme & CANINE & 68.19 & 27.33 & 58.07 & 22.65 & 61.58 & 33.53 & 26.79 & 68.78 & \textbf{55.37} & 80.07 & \textbf{57.33} & 29.87 & & 49.13 & 19.86 \\
&phoneme & XPhoneBERT(baseline) & \textbf{75.26} & 31.86 & 61.17 & 44.85 & 52.58 & \textbf{40.73} & \textbf{59.42} & 68.68 & 49.95 & 77.61 & 52.95 & 47.28 & & 55.20 & 13.83  \\
& phoneme & CONLIPA(ours) & 74.11 & \cellcolor{LightBlue}\textbf{39.95} & 60.97 & \cellcolor{LightBlue}\textbf{50.14} & \cellcolor{LightBlue}54.03 & 40.1 & 53.49 & \cellcolor{LightBlue}70.73 & \cellcolor{LightBlue}53.17 & 72.72 & 52 & \cellcolor{LightBlue}\textbf{48.44} & & \cellcolor{LightBlue}\textbf{55.82} & \cellcolor{LightBlue}\textbf{11.62}  \\
\midrule
\multirow{4}{*}{CASE 3}& & & tgk & yor & mar & jav & urd & msa & ceb & hrv & mal & tel & uzb & pan & kir &  &  \\
\cmidrule{2-18}
 & grapheme & mBERT & \textbf{74.10} & \textbf{56.60} & \textbf{74.30} & \textbf{73.59} & \textbf{57.09} & 74.98 & 64.44 & \textbf{84.93} & \textbf{69.94} & \textbf{67.24} & \textbf{80.04} & \textbf{53.98} & \textbf{68.14} & \textbf{69.18} & \textbf{9.28} \\
& grapheme & CANINE & 62.12 & 51.15 & 44.28 & 61.11 & 42.41 & \textbf{76.82} & \textbf{70.36} & 77.51 & 48.29 & 37.29 & 72.54 & 45.74 & 57.73 & 57.49 & 13.77 \\
&phoneme & XPhoneBERT(baseline) & 48.93 & 50.87 & 35.12 & 45.98 & 33.37 & 61.76 & 58.72 & 58.76 & 32.52 & 28.93 & 60.92 & 43.85 & 35.95 & 45.82 & 11.85 \\
&phoneme & CONLIPA(ours) & 48.19 & 50.05 & \cellcolor{LightBlue}38.97 & \cellcolor{LightBlue}46.24 & 31.35 & \cellcolor{LightBlue}62.83 & 58.16 & \cellcolor{LightBlue}59.17 & \cellcolor{LightBlue}39.5 & \cellcolor{LightBlue}32.57 & 60.38 & \cellcolor{LightBlue}49.48 & \cellcolor{LightBlue}39.37 & \cellcolor{LightBlue}47.40 & \cellcolor{LightBlue}10.61 \\
\bottomrule
\end{tabular}
}
\caption{\label{tab:main-table-all}
Zero-shot F1 score (\%) result in \textbf{Case 1}, \textbf{2}, and \textbf{3}. The \colorbox{LightBlue}{skyblue boxes} indicate better performance compared to the baseline, and the \textbf{bold} text represents the best performance for each case and language.
\vspace{-1em}
}
\end{table*}

\subsection{Models}
We followed \cite{sohn-etal-2024-zero} experimental setting for the three baseline models, mBERT \cite{devlin-etal-2019-bert}, CANINE \cite{clark2022canine} and XPhonebert \cite{thenguyen23_interspeech}. We also compared \cite{sohn-etal-2024-zero}'s result with ours.

We conducted experiments with models of BERT-base scale: mBERT with 177M parameters, CANINE-C with 132M, and XPhoneBERT with 87,559,687 parameters. Our model initially shares the same number of parameters as the base XPhoneBERT model, as used during pre-training with the WikiANN NER dataset following \cite{sohn-etal-2024-zero}. However, during fine-tuning on the CONLIPA dataset, we incorporated a LoRA adapter and a projection layer. The LoRA adapter adds 1,327,104 parameters, while the projection layer contributes 49,216 parameters, resulting in a total of 88,936,007 parameters. It is important to note that during zero-shot inference, the projection layer is removed, leaving only the LoRA adapter. Given the relatively small size of the LoRA adapter compared to the original XPhoneBERT parameters, this modification resulted in a substantial performance improvement with only a modest increase in model size.

\begin{table*}[t!]
\centering
\resizebox{\textwidth}{!}{
\begin{tabular}{ll|llllllllll|l}
\toprule
\multirow{2}{*}{Language}& \multirow{2}{*}{Model} & \multicolumn{10}{c}{Sample Index} & \\ \cmidrule{3-13}
 & & 1 & 2 & 3 & 4 & 5 & 6 & 7 & 8 & 9 & 10 & avg \\
\midrule
\multirow{2}{*}{eng-ori} & \cite{sohn-etal-2024-zero} & 82.41 & 60.54 & 97.17 & \textbf{74.43} & \textbf{88.50} & 90.70 & 86.91 & \textbf{93.74} & 90.46 & 81.86 & 84.67 \\
 & ours & \textbf{90.13} & \textbf{62.06} & 97.17 & 73.02 & 88.00 & \textbf{92.70} & \textbf{90.72} & 93.20 & \textbf{90.78} & \textbf{86.09} & \textbf{86.39} \\
\midrule
\multirow{2}{*}{eng-khm} & \cite{sohn-etal-2024-zero} & 80.72 & 89.65 & 88.91 & \textbf{93.94} & \textbf{98.24} & \textbf{92.32} & 57.05 & 76.12 & 69.02 & 80.05 & 82.60 \\
 & ours & \textbf{86.64} & \textbf{89.80} & \textbf{90.83} & 93.67 & 97.77 & 92.03 & \textbf{65.39} & \textbf{78.38} & \textbf{70.54} & \textbf{80.54} & \textbf{84.56} \\
\bottomrule
\end{tabular}}
\caption{\label{tab:cosine_similarity}
Cosine similarity scores(\%) for 10 samples of eng-ori and eng-khm pairs.
}
\end{table*}

\subsection{Dataset}
For training, we followed the procedure outlined in \citet{sohn-etal-2024-zero} to train XPhoneBERT on the English WikiANN NER dataset\cite{pan-etal-2017-cross}, which includes seven named entity tags: B-PER, I-PER, B-ORG, I-ORG, B-LOC, I-LOC, and O.

We then fine-tuned the model using our CONLIPA dataset with our IPA contrastive learning methodology. For zero-shot inference, we adopted the settings from \cite{sohn-etal-2024-zero} for cases 1, 2, and 3. Case 1 includes languages that were not part of the pre-training corpora of mBERT, CANINE, or XPhoneBERT. Case 2 includes only the languages that XPhoneBERT was pre-trained on, while Case 3 includes only the languages that mBERT and CANINE were pre-trained on.

\subsection{Implementation Details}
For pre-training, we followed the previous approach by setting the max sequence length, as well as both the train and validation batch sizes, to 128. We fine-tuned the pre-trained XPhoneBERT \cite{thenguyen23_interspeech} from Hugging Face \cite{wolf2019huggingface} on the English WikiAnn \cite{pan-etal-2017-cross} dataset. The training used a learning rate of 1e-5, a weight decay of 0.01, and a warmup ratio of 0.0025.

After obtaining the pre-trained checkpoint, we further performed IPA contrastive learning on our CONLIPA dataset. During this process, we froze the parameters of the original model and added a LoRA adapter with r=8, lora\_alpha=32, and lora\_dropout=0.1. Additionally, we added a linear projection layer with 64 dimensions, and only the LoRA adapter and projection layer were activated for fine-tuning for 2 epochs. All the other hyperparameters were kept the same as in the pre-training phase.

\section{Result and Analysis}
\subsection{Overall Results}
Table \ref{tab:main-table-all} compares the overall performance between our method and previous zero-shot NER approaches. It can be observed that our method outperforms the previous phonemic approach
\cite{sohn-etal-2024-zero} in all cases (Case 1, 2, and 3). Additionally, in Case 1, the most stringent zero-shot setting, our model outperformed mBERT \cite{devlin-etal-2019-bert} and CANINE \cite{clark2022canine} on average.

Notably, in Case 1, which represents a strict zero-shot setting not including any languages used in pre-training, our method shows improved performance in most languages compared to \cite{sohn-etal-2024-zero}. The average performance increases by 1.84\%, and the standard deviation decreases in the phonemic contrastive learning condition, indicating more stable and cross-lingually robust results.

\begin{table*}[thb!]
\centering
\resizebox{0.7\textwidth}{!}{
\begin{tabular}{c|cccccccc|cc}
\toprule
Model & \multicolumn{8}{c|}{Languages} & \multirow{2}{*}{AVG} & \multirow{2}{*}{STD} \\
\cmidrule{2-9}
&sin & som & mri & quy & uig & aii & kin & ilo & & \\ \midrule
XPhoneBERT & 43.61 & \textbf{38.91} & 38.07 & 51.90 & 44.82 & 31.03 & 49.67 & 73.05 & 46.38 & 12.67 \\
\midrule
$\text{Korean-16}^{\dagger}$ & \cellcolor{LightBlue}44.62 & 38.89 & \cellcolor{LightBlue}38.19 & \cellcolor{LightBlue}53.59 & \cellcolor{LightBlue}45.13 & \cellcolor{LightBlue}31.87 & 49.59 & 72.39 & \cellcolor{LightBlue}46.78 & \cellcolor{LightBlue}12.40 \\
$\text{Korean-32}^{\dagger}$ & \cellcolor{LightBlue}44.68 & 38.82 & 38.02 & \cellcolor{LightBlue}55.24 & \cellcolor{LightBlue}45.08 & 30.70 & \cellcolor{LightBlue}49.88 & \cellcolor{LightBlue}73.10 & \cellcolor{LightBlue}46.94 & 12.98 \\
$\text{Korean-64}^{\dagger}$ & \cellcolor{LightBlue}\textbf{45.90} & 38.10 & \cellcolor{LightBlue}38.60 & \cellcolor{LightBlue}55.69 & 44.44 & \cellcolor{LightBlue}33.67 & 48.25 & 72.53 & \cellcolor{LightBlue}47.15 & \cellcolor{LightBlue}12.33 \\
$\text{Korean-128}^{\dagger}$ & \cellcolor{LightBlue}45.56 & 38.49 & \cellcolor{LightBlue}38.94 & \cellcolor{LightBlue}54.28 & 44.48 & \cellcolor{LightBlue}32.60 & 47.92 & 72.22 & \cellcolor{LightBlue}46.81 & \cellcolor{LightBlue}12.21 \\
$\text{Korean-256}^{\dagger}$ & \cellcolor{LightBlue}45.88 & 37.53 & \cellcolor{LightBlue}38.73 & \cellcolor{LightBlue}54.47 & 44.38 & \cellcolor{LightBlue}33.93 & 47.70 & 72.32 & \cellcolor{LightBlue}46.87 & \cellcolor{LightBlue}12.16 \\
$\text{Korean-512}^{\dagger}$ & \cellcolor{LightBlue}45.69 & 38.70 & \cellcolor{LightBlue}39.67 & \cellcolor{LightBlue}\textbf{57.70} & \cellcolor{LightBlue}\textbf{45.17} & \cellcolor{LightBlue}34.92 & \cellcolor{LightBlue}\textbf{50.58} & \cellcolor{LightBlue}\textbf{73.35} & \cellcolor{LightBlue}\textbf{48.22} & \cellcolor{LightBlue}12.44 \\
$\text{Korean-1024}^{\dagger}$ & \cellcolor{LightBlue}45.50 & 36.77 & \cellcolor{LightBlue}40.91 & 50.40 & 42.48 & \cellcolor{LightBlue}\textbf{39.74} & 48.62 & 72.05 & \cellcolor{LightBlue}47.06 & \cellcolor{LightBlue}\textbf{11.08} \\
$\text{Korean-2048}^{\dagger}$ & \cellcolor{LightBlue}45.52 & 37.14 & \cellcolor{LightBlue}\textbf{41.36} & \cellcolor{LightBlue}54.63 & 42.14 & \cellcolor{LightBlue}37.46 & 48.93 & 72.93 & \cellcolor{LightBlue}47.51 & \cellcolor{LightBlue}11.82 \\
$\text{Korean-4096}^{\dagger}$ & \cellcolor{LightBlue}44.04 & 33.61 & \cellcolor{LightBlue}40.14 & 47.02 & 40.96 & \cellcolor{LightBlue}39.83 & 45.62 & 70.88 & 45.26 & \cellcolor{LightBlue}11.15 \\
$\text{Korean-7521}^{\dagger}$ & 32.52 & 25.66 & 28.52 & 42.20 & 36.80 & \cellcolor{LightBlue}32.16 & 43.61 & 64.15 & 38.20 & \cellcolor{LightBlue}12.19 \\
\midrule
\bottomrule
\end{tabular}
}
\caption{\label{tab:korean_num_case1}
Ablation study on Korean data number in Case 1. $\dagger$ indicates that the model was trained using all 10 languages of CONLIPA, but with a different number of samples of Korean. The \colorbox{LightBlue}{skyblue boxes} indicate better performance compared to the baseline, and the \textbf{bold} text represents the best performance for each case and language.}
\end{table*}

\subsection{Cosine Similarity of Phonemic Representation}
\label{cosine_similarity}
The goal of IPA contrastive learning is to align the cross-lingual representations of languages with similar pronunciations but slightly different IPA transcriptions. To evaluate this, we computed the distance between named entity pairs in English and two low-resource languages, Oriya and Khmer, where each pair has the same meaning, similar pronunciation but different IPA transcription. The distance between the embeddings from each language was calculated using the cosine similarity metric. Figures \ref{fig:eng_oriya_pair} and \ref{fig:eng_khmer_pair} in the Appendix present the 10 samples for Oriya and Khmer, respectively.

As shown in Table \ref{tab:cosine_similarity}, compared to \cite{sohn-etal-2024-zero}, the results after applying our IPA contrastive learning on both eng-ori and eng-khm showed higher cosine similarity scores in most cases, with the average score also being higher for our method. This demonstrates that our method successfully brought phonemic embeddings with similar meanings and pronunciations closer together across different languages. The t-SNE visualization results for these samples are also provided in the section \ref{t_sne_visualization} of Appendix.

\begin{table*}[thb!]
\centering
\resizebox{0.7\textwidth}{!}{
\begin{tabular}{c|cccccccc|cc}
\toprule
Model & \multicolumn{8}{c|}{Languages} & \multirow{2}{*}{AVG} & \multirow{2}{*}{STD} \\
\cmidrule{2-9}
&sin & som & mri & quy & uig & aii & kin & ilo & & \\ \midrule
XPhoneBERT & 43.61 & \textbf{38.91} & 38.07 & 51.90 & 44.82 & 31.03 & 49.67 & 73.05 & 46.38 & 12.67 \\
\midrule
Swahili & \cellcolor{LightBlue}44.74 & 38.71 & \cellcolor{LightBlue}38.12 & \cellcolor{LightBlue}53.66 & \cellcolor{LightBlue}44.89 & \cellcolor{LightBlue}31.65 & 49.43 & \cellcolor{LightBlue}73.24 & \cellcolor{LightBlue}46.81 & 12.71 \\
Indonesian & \cellcolor{LightBlue}44.43 & \cellcolor{LightBlue}39.05 & \cellcolor{LightBlue}39.00 & \cellcolor{LightBlue}55.53 & \cellcolor{LightBlue}44.84 & \cellcolor{LightBlue}32.54 & 49.43 & 72.46 & \cellcolor{LightBlue}47.16 & \cellcolor{LightBlue}12.39 \\
Hindi & \cellcolor{LightBlue}44.62 & 38.53 & \cellcolor{LightBlue}38.08 & \cellcolor{LightBlue}53.69 & \cellcolor{LightBlue}44.97 & 30.98 & 49.28 & \cellcolor{LightBlue}73.25 & \cellcolor{LightBlue}46.68 & 12.85 \\
Mandarin & \cellcolor{LightBlue}44.37 & \cellcolor{LightBlue}\textbf{39.2} & \cellcolor{LightBlue}38.56 & \cellcolor{LightBlue}53.61 & \cellcolor{LightBlue}45.00 & \cellcolor{LightBlue}31.28 & 49.63 & 72.66 & \cellcolor{LightBlue}46.79 & \cellcolor{LightBlue}12.53 \\
Arabic & \cellcolor{LightBlue}44.46 & \cellcolor{LightBlue}39.11 & \cellcolor{LightBlue}38.55 & \cellcolor{LightBlue}55.02 & \cellcolor{LightBlue}44.90 & \cellcolor{LightBlue}32.56 & 49.4 & 72.71 & \cellcolor{LightBlue}47.09 & \cellcolor{LightBlue}12.44 \\
Vietnamese & \cellcolor{LightBlue}44.53 & \cellcolor{LightBlue}39.07 & 38.03 & \cellcolor{LightBlue}55.31 & \cellcolor{LightBlue}44.95 & \cellcolor{LightBlue}31.94 & \cellcolor{LightBlue}50.1 & 72.69 & \cellcolor{LightBlue}47.08 & \cellcolor{LightBlue}12.64 \\
Thai & \cellcolor{LightBlue}44.61 & \cellcolor{LightBlue}39.15 & 37.94 & \cellcolor{LightBlue}54.53 & \cellcolor{LightBlue}\textbf{45.25} & \cellcolor{LightBlue}31.94 & \cellcolor{LightBlue}49.89 & 72.42 & \cellcolor{LightBlue}46.97 & \cellcolor{LightBlue}12.48 \\
Tamil & \cellcolor{LightBlue}44.43 & \cellcolor{LightBlue}39.07 & 37.95 & \cellcolor{LightBlue}54.68 & \cellcolor{LightBlue}45.00 & 30.75 & \cellcolor{LightBlue}50.01 & 72.81 & \cellcolor{LightBlue}46.84 & \cellcolor{LightBlue}12.01 \\
Turkish & \cellcolor{LightBlue}44.62 & 38.89 & \cellcolor{LightBlue}38.22 & \cellcolor{LightBlue}54.93 & \cellcolor{LightBlue}44.98 & 30.81 & \cellcolor{LightBlue}50.09 & \cellcolor{LightBlue}73.24 & \cellcolor{LightBlue}46.97 & 12.96 \\
Korean & \cellcolor{LightBlue}44.57 & 38.51 & \cellcolor{LightBlue}38.75 & \cellcolor{LightBlue}55.48 & \cellcolor{LightBlue}44.86 & \cellcolor{LightBlue}33.56 & 49.5 & 72.5 & \cellcolor{LightBlue}47.22 & \cellcolor{LightBlue}12.30 \\
Total & \cellcolor{LightBlue}\textbf{45.69} & 38.7 & \cellcolor{LightBlue}\textbf{39.67} & \cellcolor{LightBlue}\textbf{57.7} & \cellcolor{LightBlue}45.17 & \cellcolor{LightBlue}\textbf{34.92} & \cellcolor{LightBlue}\textbf{50.58} & \cellcolor{LightBlue}\textbf{73.35} & \cellcolor{LightBlue}\textbf{48.22} & \cellcolor{LightBlue}12.44  \\
\midrule
\bottomrule
\end{tabular}
}
\caption{\label{tab:single-table-case1}
Ablation study on each language in case 1. The \colorbox{LightBlue}{skyblue boxes} indicate better performance compared to the baseline, and the \textbf{bold} text represents the best performance for each inference language.}
\end{table*}

\subsection{Ablation Study}
\subsubsection{Ablation on the number of Korean samples}
\label{sec:korean_data_num_ablation}
As can be seen in Table \ref{tab:language_family}, the number of Korean data samples is 7,521, which is significantly higher than that of the other languages. To determine the optimal number of samples for achieving the best performance, we conducted an ablation study by varying the amount of Korean data used in training the model with IPA contrastive learning.

We conducted experiments by gradually increasing the number of Korean data samples, doubling them from 16, 32, 64, ..., up to 7,521, while keeping the data samples of the other 10 languages fixed. As shown in Table \ref{tab:korean_num_case1}, the best performance was achieved when the number of Korean data samples was 512. This demonstrates that simply increasing the number of data samples used for IPA contrastive learning does not always lead to better results. While IPA contrastive learning helps bring the representations of similar-sounding words across different languages closer together, excessive usage of it may potentially harm the representations of models pre-trained on original NER datasets. The experimental results of Case 2,3 are also available on Table \ref{tab:korean_data_num} of Appendix.

\subsubsection{Ablation on each Language}
We conducted experiments using only the data from each of the 10 languages in CONLIPA to identify which language performs best when training the model with IPA contrastive learning. As shown in Table \ref{tab:single-table-case1}, Korean achieved the best performance, followed by Indonesian, Arabic, and Vietnamese. However, we can still observe that the \emph{Total} result, using all 10 languages, performed the best, indicating that the data from multiple languages are complementary to each other. The experimental results for Case 2 and Case 3 can also be found in Appendix Table \ref{tab:single-table-all}.

\section{Conclusion}
This paper proposes a novel cross-lingual \textbf{IPA} \textbf{C}ontrastive learning(\textbf{IPAC}) methodology to make the phonemic representations of languages with similar pronunciations more similar, aimed at zero-shot cross-lingual NER for low-resource languages. For this purpose, we selected 10 commonly used language families and introduce the \textbf{CON}trastive \textbf{L}earning with \textbf{IPA}(\textbf{CONLIPA}) dataset, which includes IPA pairs of similar-sounding words between English and these languages.

Through experiments, we demonstrate that our approach outperforms existing subword, character grapheme-based models, and the basic phoneme-based model. Performance improvements across all cases 1, 2, and 3 confirm the our method's effect on the cross-lingual generalization of  phonemic representation, which is crucial for zero-shot NER tasks in low-resource languages where data is scarce.

\section{Limitations}
Our methodology does not consider all language families worldwide, but rather focuses on 10 language families. Additionally, it is difficult to claim that the representative language selected from each of the 10 language families fully represents all the characteristics of every language within that family. However, our approach demonstrates the potential to improve performance for low-resource languages by leveraging data from high-resource languages, which are relatively easier to obtain.

\section{Ethics Statement}
In this study, we utilize the publicly available WikiANN dataset \cite{pan-etal-2017-cross} to train various models across different languages, ensuring that no ethical concerns arise. During the creation of the CONLIPA dataset, we encountered no ethical issues related to its curation or annotation. There were no significant ethical concerns, such as violent or offensive content, and the dataset was used in accordance with its intended purpose.

\bibliography{acl_latex}

\input{appendix}

\end{document}

%% file: appendix.tex
\appendix

\section{Language Codes}
Table \ref{tab:lang_code} presents the ISO 639-3 language codes for all the languages utilized in the experiments.

\begin{table}
\centering
\resizebox{0.7\columnwidth}{!}{
\begin{tabular}{ll}
\toprule
Language & ISO 639-3 \\
\midrule
Amharic & amh \\
Assyrian Neo-Aramaic & aii\\ 
Ayacucho quechua & quy\\ 
Cebuano & ceb\\
Croatian & hrv\\
English & eng\\
Esperanto & epo\\
Ilocano & ilo\\
Javanese & jav\\
Khmer & khm\\
Kinyarwanda & kin \\
Kyrgyz & kir \\
Malay & msa \\
Malayalam & mal \\
Maltese & mlt \\
Maori & mri \\
Marathi & mar \\
Punjabi & pan \\
Sinhala & sin \\
Somali & som \\
Tajik & tgk \\
Telugu & tel \\
Turkmen & tuk \\
Urdu & urd \\
Uyghur & uig \\
Uzbek & uzb \\
Yoruba & yor \\
Swahili & swa \\
Indonesian & ind \\
Hindi & hin \\
Mandarin & cmn \\
Arabic & ara \\
Vietnamese & vie \\
Thai & tha \\
Tamil & tam \\
Turkish & tur \\
Korean & kor \\
\bottomrule
\end{tabular}}
\caption{\label{tab:lang_code}
Language codes for all languages used in the experiments.
}
\end{table}

\section{Benchmark and License}
Table \ref{tab:dataset_stats} provides information on the datasets, including their statistics and licensing details. Additionally, the CharsiuG2P toolkit~\cite{Zhu2022ByT5MF}, used for transliteration, is employed under the MIT license.

\begin{table}[]
    \centering
    \scriptsize
    \begin{tabular}{ccccccc}
    \toprule
         Dataset & Lang. & Script & Train & Dev & Test & License \\ \midrule
         \multirow{44}{*}{WikiANN} & \texttt{eng}& Latn & 20k & 10k & 10k & \multirow{44}{*}{ODC-BY} \\
                                & \texttt{sin}& Sinh & 100 & 100 & 100 & \\
                                & \texttt{som}& Latn & 100 & 100 & 100 & \\
                                & \texttt{mri}& Latn & 100 & 100 & 100 & \\
                                & \texttt{quy}& Latn & 100 & 100 & 100 & \\
                                & \texttt{uig}& Arab & 100 & 100 & 100 & \\
                                & \texttt{aii}& Syrc & 100 & 100 & 100 & \\
                                & \texttt{kin}& Latn & 100 & 100 & 100 & \\
                                & \texttt{ilo}& Latn & 100 & 100 & 100 & \\
                                & \texttt{epo}& Latn & 15k & 10k & 10k & \\
                                & \texttt{khm}& Khmr & 100 & 100 & 100 & \\
                                & \texttt{tuk}& Latn & 100 & 100 & 100 & \\
                                & \texttt{amh}& Ethi & 100 & 100 & 100 & \\
                                & \texttt{mlt}& Latn & 100 & 100 & 100 & \\
                                & \texttt{ori}& Orya & 100 & 100 & 100 & \\
                                & \texttt{san}& Deva & 100 & 100 & 100 &  \\
                                & \texttt{ina}& Latn & 100 & 100 & 100 &  \\
                                & \texttt{grn}& Latn & 100 & 100 & 100 & \\
                                & \texttt{bel}& Cyrl & 15k & 1k & 1k & \\
                                & \texttt{kur}& Latn & 100 & 100 & 100 & \\
                                & \texttt{snd}& Arab & 100 & 100 & 100 & \\
                                & \texttt{tgk}& Cyrl & 100 & 100 & 100 & \\
                                & \texttt{yor}& Latn & 100 & 100 & 100 &  \\
                                & \texttt{mar}& Deva & 5k & 1k & 1k &  \\
                                & \texttt{jav}& Latn & 100 & 100 & 100 & \\
                                & \texttt{urd}& Arab & 20k & 1k & 1k & \\
                                & \texttt{msa}& Latn & 20k & 1k & 1k & \\
                                & \texttt{ceb}& Latn & 100 & 100 & 100 & \\
                                & \texttt{hrv}& Latn & 20k & 10k & 10k & \\
                                & \texttt{mal}& Mlym & 10k & 1k & 1k &  \\
                                & \texttt{tel}& Telu & 1k & 1k & 1k &  \\
                                & \texttt{uzb}& Cyrl & 1k & 1k & 1k & \\
                                & \texttt{pan}& Guru & 100 & 100 & 100 & \\
                                & \texttt{kir}& Latn & 100 & 100 & 100 & \\
                                & \texttt{swa}& Latn & 1k & 1k & 1k & \\
                                & \texttt{ind}& Latn & 20k & 10k & 10k & \\
                                & \texttt{hin}& Deva & 5k & 1k & 1k & \\
                                & \texttt{cmn}& Han & 20k & 10k & 10k & \\
                                & \texttt{ara}& Arab & 20k & 10k & 10k & \\
                                & \texttt{vie}& Latn & 20k & 10k & 10k & \\
                                & \texttt{tha}& Thai & 20k & 10k & 10k & \\
                                & \texttt{tam}& Telu & 15k & 1k & 1k & \\
                                & \texttt{tur}& Latn & 20k & 10k & 10k & \\
                                & \texttt{kor}& Hangul & 20k & 10k & 10k & \\
         \bottomrule
  \end{tabular}
  \caption{Statistics and license types for the dataset. The table lists the script, number of examples in the training, development, and testing sets for languages in the WikiANN dataset. The dataset is strictly used within the bounds of these licenses.}
      \label{tab:dataset_stats}
\end{table}

\section{Experimental Result on the trained High Resource Language}

\begin{table*}[thb!]
\centering
\resizebox{0.9\textwidth}{!}{
\begin{tabular}{c|ccccccccccc|cc}
\toprule
\multirow{2}{*}{Train Language} & \multicolumn{11}{c|}{Zero-shot Inference Language} & \multirow{2}{*}{AVG} & \multirow{2}{*}{STD} \\
\cmidrule{2-12}
& eng & kor & swa & ind & hin & cmn & ara & vie & tha & tam & tur \\ \midrule
XPhoneBERT & 76.78 & 54.88 & 61.40 & 64.00 & 64.99 & 39.16 & 55.24 & 58.49 & 16.74 & 59.39 & 67.28 & 56.21 & 16.05 \\
\midrule
kor & \cellcolor{LightBlue}76.69 & \cellcolor{LightBlue}55.09 & 61.36 & \cellcolor{LightBlue}64.02 & \cellcolor{LightBlue}65.26 & 38.83 & 55.22 & \cellcolor{LightBlue}\textbf{58.64} & 16.69 & \cellcolor{LightBlue}59.96 & 67.18 & \cellcolor{LightBlue}56.27 & 16.11  \\
swa & 76.79 & 54.81 & 61.36 & \cellcolor{LightBlue}64.07 & 64.81 & \cellcolor{LightBlue}39.20 & 55.14 & 58.46 & \cellcolor{LightBlue}16.75 & 59.28 & 67.18 & 56.17 & \cellcolor{LightBlue}\textbf{16.03} \\
ind & \cellcolor{LightBlue}76.8 & \cellcolor{LightBlue}55.03 & 61.23 & 63.98 & \cellcolor{LightBlue}65.08 & \cellcolor{LightBlue}\textbf{39.22} & 55.22 & \cellcolor{LightBlue}58.53 & \cellcolor{LightBlue}16.77 & \cellcolor{LightBlue}59.54 & \cellcolor{LightBlue}67.34 & \cellcolor{LightBlue}56.25 & \cellcolor{LightBlue}16.05 \\
hin & 76.74 & 54.68 & \cellcolor{LightBlue}61.48 & \cellcolor{LightBlue}64.04 & 64.86 & \cellcolor{LightBlue}\textbf{39.22} & 55.10 & \cellcolor{LightBlue}58.51 & 16.72 & \cellcolor{LightBlue}59.46 & 67.15 & 56.18 & \cellcolor{LightBlue}16.04 \\
cmn & \cellcolor{LightBlue}76.85 & \cellcolor{LightBlue}55.03 & 61.31 & 63.95 & \cellcolor{LightBlue}65.07 & 39.14 & \cellcolor{LightBlue}55.26 & 58.45 & \cellcolor{LightBlue}16.76 & \cellcolor{LightBlue}59.51 & \cellcolor{LightBlue}67.35 & \cellcolor{LightBlue}56.24 & 16.06 \\
ara & \cellcolor{LightBlue}76.82 & \cellcolor{LightBlue}55.00 & 61.34 & \cellcolor{LightBlue}64.08 & \cellcolor{LightBlue}65.13 & \cellcolor{LightBlue}39.19 & 55.15 & \cellcolor{LightBlue}58.57 & 16.74 & \cellcolor{LightBlue}59.55 & 67.28 & \cellcolor{LightBlue}56.26 & 16.07 \\
vie & \cellcolor{LightBlue}76.87 & \cellcolor{LightBlue}55.11 & \cellcolor{LightBlue}\textbf{61.52} & \cellcolor{LightBlue}64.03 & \cellcolor{LightBlue}65.21 & 39.06 & \cellcolor{LightBlue}55.31 & \cellcolor{LightBlue}58.52 & \cellcolor{LightBlue}16.75 & \cellcolor{LightBlue}59.66 & \cellcolor{LightBlue}67.36 & \cellcolor{LightBlue}56.31 & 16.10 \\
tha & \cellcolor{LightBlue}76.87 & \cellcolor{LightBlue}55.08 & 61.33 & 63.99 & \cellcolor{LightBlue}65.25 & 39.07 & \cellcolor{LightBlue}\textbf{55.32} & \cellcolor{LightBlue}58.51 & \cellcolor{LightBlue}16.77 & \cellcolor{LightBlue}59.49 & \cellcolor{LightBlue}67.43 & \cellcolor{LightBlue}56.28 & 16.09 \\
tam & \cellcolor{LightBlue}76.82 & \cellcolor{LightBlue}55.01 & 61.40 & 63.96 & \cellcolor{LightBlue}65.01 & 39.14 & \cellcolor{LightBlue}55.27 & 58.43 & \cellcolor{LightBlue}16.76 & 59.38 & \cellcolor{LightBlue}67.36 & \cellcolor{LightBlue}56.23 & 16.06 \\
tur & 76.77 & 54.73 & \cellcolor{LightBlue}61.45 & \cellcolor{LightBlue}64.02 & 64.84 & \cellcolor{LightBlue}39.19 & 55.08 & 58.42 & \cellcolor{LightBlue}16.75 & \cellcolor{LightBlue}59.44 & 67.23 & 56.17 & \cellcolor{LightBlue}16.04 \\
total & \cellcolor{LightBlue}\textbf{76.93} & \cellcolor{LightBlue}\textbf{55.82} & 61.15 & \cellcolor{LightBlue}\textbf{64.09} & \cellcolor{LightBlue}\textbf{66.13} & 38.87 & \cellcolor{LightBlue}55.25 & 58.49 & \cellcolor{LightBlue}\textbf{16.87} & \cellcolor{LightBlue}\textbf{60.49} & \cellcolor{LightBlue}\textbf{67.55} & \cellcolor{LightBlue}\textbf{56.51} & 16.17  \\
\bottomrule
\end{tabular}
}
\caption{\label{tab:high_resource_validation}
F1 score(\%) for zero-shot inference on each high-resource language after training on each language of CONLIPA.
}
\end{table*}

The main task of our paper is to perform NER in a strict zero-shot setting, where the inference is conducted on a low-resource language that has never been seen before. However, we also compared the validation set results before and after training on the CONLIPA dataset, which consists of 10 high-resource languages used for Cross-lingual IPA contrastive learning.

As shown in Table \ref{tab:high_resource_validation}, in most cases, the performance improved compared to the existing baseline. Although there were occasional instances where the performance dropped below the baseline, the maximum performance improvement was 1.14, while the maximum performance degradation was 0.33. Since the largest performance drop is small, it suggests that performing IPA contrastive learning using the CONLIPA dataset may also be effective in improving the performance of high-resource languages. Additionally, it can be observed that using all 10 languages as \emph{total} shows the best performance both on average and for most individual languages in high-resource languages, too. This suggests that the interaction among 10 representative languages from 10 different language families leads to better results.

\section{Experimental Result with number of Korean data}
We present the ablation study examining the number of Korean instances across all three cases in Table \ref{tab:korean_data_num}. 

\begin{table*}
\centering
\resizebox{\textwidth}{!}{
\begin{tabular}{c|cccccccccccccccc}
\toprule
Case & Model & \multicolumn{13}{c}{Languages} & AVG & STD \\
\midrule
\multirow{4}{*}{CASE 1}&&sin & som & mri & quy & uig & aii & kin & ilo & & & & & & & \\ \cmidrule{2-17}
& XPhoneBERT & 43.61 & \textbf{38.91} & 38.07 & 51.90 & 44.82 & 31.03 & 49.67 & 73.05 & & & & & & 46.38 & 12.67  \\
\cmidrule{2-17}
& $\text{Korean-16}^{\dagger}$ & \cellcolor{LightBlue}44.62 & 38.89 & \cellcolor{LightBlue}38.19 & \cellcolor{LightBlue}53.59 & \cellcolor{LightBlue}45.13 & \cellcolor{LightBlue}31.87 & 49.59 & 72.39 &&&&&& \cellcolor{LightBlue}46.78 & \cellcolor{LightBlue}12.40 \\
& $\text{Korean-32}^{\dagger}$ & \cellcolor{LightBlue}44.68 & 38.82 & 38.02 & \cellcolor{LightBlue}55.24 & \cellcolor{LightBlue}45.08 & 30.70 & \cellcolor{LightBlue}49.88 & \cellcolor{LightBlue}73.10 &&&&&& \cellcolor{LightBlue}46.94 & 12.98 \\
& $\text{Korean-64}^{\dagger}$ & \cellcolor{LightBlue}\textbf{45.90} & 38.10 & \cellcolor{LightBlue}38.60 & \cellcolor{LightBlue}55.69 & 44.44 & \cellcolor{LightBlue}33.67 & 48.25 & 72.53 &&&&&& \cellcolor{LightBlue}47.15 & \cellcolor{LightBlue}12.33 \\
& $\text{Korean-128}^{\dagger}$ & \cellcolor{LightBlue}45.56 & 38.49 & \cellcolor{LightBlue}38.94 & \cellcolor{LightBlue}54.28 & 44.48 & \cellcolor{LightBlue}32.60 & 47.92 & 72.22 &&&&&& \cellcolor{LightBlue}46.81 & \cellcolor{LightBlue}12.21 \\
& $\text{Korean-256}^{\dagger}$ & \cellcolor{LightBlue}45.88 & 37.53 & \cellcolor{LightBlue}38.73 & \cellcolor{LightBlue}54.47 & 44.38 & \cellcolor{LightBlue}33.93 & 47.70 & 72.32 &&&&&& \cellcolor{LightBlue}46.87 & \cellcolor{LightBlue}12.16 \\
& $\text{Korean-512}^{\dagger}$ & \cellcolor{LightBlue}45.69 & 38.70 & \cellcolor{LightBlue}39.67 & \cellcolor{LightBlue}\textbf{57.70} & \cellcolor{LightBlue}\textbf{45.17} & \cellcolor{LightBlue}34.92 & \cellcolor{LightBlue}\textbf{50.58} & \cellcolor{LightBlue}\textbf{73.35} &&&&&& \cellcolor{LightBlue}\textbf{48.22} & \cellcolor{LightBlue}12.44 \\
& $\text{Korean-1024}^{\dagger}$ & \cellcolor{LightBlue}45.50 & 36.77 & \cellcolor{LightBlue}40.91 & 50.40 & 42.48 & \cellcolor{LightBlue}\textbf{39.74} & 48.62 & 72.05 &&&&&& \cellcolor{LightBlue}47.06 & \cellcolor{LightBlue}\textbf{11.08} \\
& $\text{Korean-2048}^{\dagger}$ & \cellcolor{LightBlue}45.52 & 37.14 & \cellcolor{LightBlue}\textbf{41.36} & \cellcolor{LightBlue}54.63 & 42.14 & \cellcolor{LightBlue}37.46 & 48.93 & 72.93 &&&&&& \cellcolor{LightBlue}47.51 & \cellcolor{LightBlue}11.82 \\
& $\text{Korean-4096}^{\dagger}$ & \cellcolor{LightBlue}44.04 & 33.61 & \cellcolor{LightBlue}40.14 & 47.02 & 40.96 & \cellcolor{LightBlue}39.83 & 45.62 & 70.88 &&&&&& 45.26 & \cellcolor{LightBlue}11.15 \\
& $\text{Korean-7521}^{\dagger}$ & 32.52 & 25.66 & 28.52 & 42.20 & 36.80 & \cellcolor{LightBlue}32.16 & 43.61 & 64.15 &&&&&& 38.20 & \cellcolor{LightBlue}12.19 \\
\midrule
\multirow{4}{*}{CASE 2} & & epo & khm & tuk & amh & mlt & ori & san & ina & grn & bel & kur & snd & &  &  \\
\cmidrule{2-17}
& XPhoneBERT & \textbf{75.26} & 31.86 & \textbf{61.17} & 44.85 & 52.58 & 40.73 & \textbf{59.42} & 68.68 & 49.95 & \textbf{77.61} & 52.95 & 47.28 & & 55.20 & 13.83  \\
\cmidrule{2-17}
& $\text{Korean-16}^{\dagger}$ & 73.48 & \cellcolor{LightBlue}38.79 & 59.45 & \cellcolor{LightBlue}\textbf{52.41} & \cellcolor{LightBlue}\textbf{55.46} & 39.91 & 54.14 &\cellcolor{LightBlue} \cellcolor{LightBlue}70.22 & \cellcolor{LightBlue}\textbf{54.98} & 72.48 & 51.89 & \cellcolor{LightBlue}48.19 && \cellcolor{LightBlue}55.95 & \cellcolor{LightBlue}11.45 \\
& $\text{Korean-32}^{\dagger}$ & 73.35 & \cellcolor{LightBlue}38.56 & 59.00 & \cellcolor{LightBlue}51.96 & \cellcolor{LightBlue}55.09 & 40.22 & 54.59 & \cellcolor{LightBlue}70.03 & \cellcolor{LightBlue}54.55 & 72.29 & 52.15 & \cellcolor{LightBlue}47.96 && \cellcolor{LightBlue}55.81 & \cellcolor{LightBlue}11.38 \\
& $\text{Korean-64}^{\dagger}$ & 73.78 & \cellcolor{LightBlue}39.74 & 59.28 & \cellcolor{LightBlue}52.17 & \cellcolor{LightBlue}53.76 & 40.23 & 53.58 & \cellcolor{LightBlue}70.37 & \cellcolor{LightBlue}53.61 & 72.32 & 51.89 & \cellcolor{LightBlue}47.88 && \cellcolor{LightBlue}55.72 & \cellcolor{LightBlue}11.39 \\
& $\text{Korean-128}^{\dagger}$ & 73.66 & \cellcolor{LightBlue}39.85 & 59.60 & \cellcolor{LightBlue}52.08 & \cellcolor{LightBlue}53.62 & 40.39 & 53.38 & \cellcolor{LightBlue}70.50 & \cellcolor{LightBlue}54.28 & 72.45 & \textbf{52.21} & \cellcolor{LightBlue}48.94 && \cellcolor{LightBlue}55.91 & \cellcolor{LightBlue}11.30 \\
& $\text{Korean-256}^{\dagger}$ & 73.79 & \cellcolor{LightBlue}40.01 & 60.13 & \cellcolor{LightBlue}52.01 & \cellcolor{LightBlue}53.56 & \cellcolor{LightBlue}41.12 & 53.68 & \cellcolor{LightBlue}70.80 & \cellcolor{LightBlue}53.36 & 72.38 & 52.14 & \cellcolor{LightBlue}48.69 && \cellcolor{LightBlue}\textbf{55.97} & \cellcolor{LightBlue}11.28 \\
& $\text{Korean-512}^{\dagger}$ & 74.11 & \cellcolor{LightBlue}39.95 & 60.97 & \cellcolor{LightBlue}50.14 & \cellcolor{LightBlue}54.03 & 40.1 & 53.49 & \cellcolor{LightBlue}70.73 & \cellcolor{LightBlue}53.17 & 72.72 & 52.00 & \cellcolor{LightBlue}48.44 && \cellcolor{LightBlue}55.82 & \cellcolor{LightBlue}11.62 \\
& $\text{Korean-1024}^{\dagger}$ & 73.96 & \cellcolor{LightBlue}\textbf{44.88} & 59.86 & \cellcolor{LightBlue}49.16 & 51.71 & \cellcolor{LightBlue}41.15 & 52.58 & \cellcolor{LightBlue}\textbf{71.05} & \cellcolor{LightBlue}51.73 & 72.80 & 51.76 & \cellcolor{LightBlue}\textbf{50.02} && \cellcolor{LightBlue}55.89 & \cellcolor{LightBlue}11.03 \\
& $\text{Korean-2048}^{\dagger}$ & 73.98 & \cellcolor{LightBlue}41.60 & 60.37 & \cellcolor{LightBlue}49.64 & 52.46 & \cellcolor{LightBlue}\textbf{41.42} & 53.53 & \cellcolor{LightBlue}70.86 & \cellcolor{LightBlue}52.23 & 72.59 & 52.13 & \cellcolor{LightBlue}48.80 && \cellcolor{LightBlue}55.80 & \cellcolor{LightBlue}11.27 \\
& $\text{Korean-4096}^{\dagger}$ & 72.81 & \cellcolor{LightBlue}42.18 & 57.96 & \cellcolor{LightBlue}50.42 & 50.78 & \cellcolor{LightBlue}43.14 & 50.46 & 68.43 & \cellcolor{LightBlue}50.13 & 71.59 & 52.33 & \cellcolor{LightBlue}49.18 && 54.95 & \cellcolor{LightBlue}\textbf{10.49} \\
& $\text{Korean-7521}^{\dagger}$ & 68.01 & \cellcolor{LightBlue}38.03 & 56.40 & \cellcolor{LightBlue}46.82 & 46.69 & 39.18 & 48.62 & 65.08 & 45.87 & 69.23 & 52.78 & 45.70 && 51.87 & \cellcolor{LightBlue}10.65 \\
\midrule
\multirow{4}{*}{CASE 3}& & tgk & yor & mar & jav & urd & msa & ceb & hrv & mal & tel & uzb & pan & kir &  &  \\
\cmidrule{2-17}
& XPhoneBERT & 48.93 & \textbf{50.87} & 35.12 & 45.98 & 33.37 & 61.76 & 58.72 & 58.76 & 32.52 & 28.93 & 60.92 & 43.85 & 35.95 & 45.82 & 11.85 \\
\cmidrule{2-17}
& $\text{Korean-16}^{\dagger}$ & \cellcolor{LightBlue}49.01 & 50.19 & \cellcolor{LightBlue}38.15 & \cellcolor{LightBlue}46.19 & 32.63 & \cellcolor{LightBlue}61.78 & \cellcolor{LightBlue}\textbf{59.21} & \cellcolor{LightBlue}58.95 & \cellcolor{LightBlue}39.52 & \cellcolor{LightBlue}32.54 & 60.70 & \cellcolor{LightBlue}49.36 & \cellcolor{LightBlue}38.49 & \cellcolor{LightBlue}\textbf{47.44} & \cellcolor{LightBlue}10.56 \\
& $\text{Korean-32}^{\dagger}$ & 48.90 & 50.60 & \cellcolor{LightBlue}37.94 & \cellcolor{LightBlue}45.99 & \cellcolor{LightBlue}\textbf{33.39} & \cellcolor{LightBlue}61.79 & 58.72 & 58.72 & \cellcolor{LightBlue}38.93 & \cellcolor{LightBlue}32.12 & \cellcolor{LightBlue}\textbf{61.08} & \cellcolor{LightBlue}47.56 & \cellcolor{LightBlue}37.73 & \cellcolor{LightBlue}47.19 & \cellcolor{LightBlue}10.60 \\
& $\text{Korean-64}^{\dagger}$ & \cellcolor{LightBlue}49.22 & 49.55 & \cellcolor{LightBlue}38.03 & \cellcolor{LightBlue}46.20 & 32.09 & \cellcolor{LightBlue}62.36 & 58.25 & 58.72 & \cellcolor{LightBlue}39.40 & \cellcolor{LightBlue}32.17 & 60.40 & \cellcolor{LightBlue}48.51 & \cellcolor{LightBlue}38.34 & \cellcolor{LightBlue}47.17 & \cellcolor{LightBlue}10.60 \\
& $\text{Korean-128}^{\dagger}$ & \cellcolor{LightBlue}\textbf{49.56} & 49.38 & \cellcolor{LightBlue}38.54 & 45.54 & 32.16 & \cellcolor{LightBlue}61.81 & \cellcolor{LightBlue}59.13 & \cellcolor{LightBlue}59.12 & \cellcolor{LightBlue}\textbf{39.83} & \cellcolor{LightBlue}32.69 & 60.33 & \cellcolor{LightBlue}48.99 & \cellcolor{LightBlue}38.81 & \cellcolor{LightBlue}47.38 & \cellcolor{LightBlue}10.50 \\
& $\text{Korean-256}^{\dagger}$ & 48.88 & 49.50 & \cellcolor{LightBlue}38.41 & \cellcolor{LightBlue}46.03 & 31.70 & \cellcolor{LightBlue}62.16 & 58.68 & \cellcolor{LightBlue}58.90 & \cellcolor{LightBlue}\textbf{39.83} & \cellcolor{LightBlue}32.50 & 60.25 & \cellcolor{LightBlue}49.42 & \cellcolor{LightBlue}38.39 & \cellcolor{LightBlue}47.28 & \cellcolor{LightBlue}10.58 \\
& $\text{Korean-512}^{\dagger}$ & 48.19 & 50.05 & \cellcolor{LightBlue}\textbf{38.97} & \cellcolor{LightBlue}46.24 & 31.35 & \cellcolor{LightBlue}\textbf{62.83} & 58.16 & \cellcolor{LightBlue}\textbf{59.17} & \cellcolor{LightBlue}39.50 & \cellcolor{LightBlue}32.57 & 60.38 & \cellcolor{LightBlue}49.48 & \cellcolor{LightBlue}\textbf{39.37} & \cellcolor{LightBlue}47.40 & \cellcolor{LightBlue}10.61 \\
& $\text{Korean-1024}^{\dagger}$ & 45.16 & 47.22 & \cellcolor{LightBlue}38.90 & \cellcolor{LightBlue}\textbf{46.63} & 28.61 & \cellcolor{LightBlue}62.57 & 58.36 & 58.26 & \cellcolor{LightBlue}39.64 & \cellcolor{LightBlue}\textbf{33.16} & 57.91 & \cellcolor{LightBlue}51.73 & \cellcolor{LightBlue}36.86 & \cellcolor{LightBlue}46.54 & \cellcolor{LightBlue}10.77 \\
& $\text{Korean-2048}^{\dagger}$ & 45.87 & 48.20 & \cellcolor{LightBlue}38.45 & 45.77 & 29.32 & \cellcolor{LightBlue}62.69 & 56.68 & 58.08 & \cellcolor{LightBlue}39.12 & \cellcolor{LightBlue}32.13 & 57.60 & \cellcolor{LightBlue}\textbf{51.77} & \cellcolor{LightBlue}38.32 & \cellcolor{LightBlue}46.46 & \cellcolor{LightBlue}10.58 \\
& $\text{Korean-4096}^{\dagger}$ & 45.18 & 45.81 & \cellcolor{LightBlue}37.48 & 45.57 & 28.17 & 61.26 & 53.28 & 56.11 & \cellcolor{LightBlue}40.01 & \cellcolor{LightBlue}31.82 & 55.50 & \cellcolor{LightBlue}52.10 & \cellcolor{LightBlue}36.12 & 45.26 & \cellcolor{LightBlue}10.14 \\
& $\text{Korean-7521}^{\dagger}$ & 31.09 & 42.08 & 35.00 & 44.58 & 23.80 & 56.53 & 50.39 & 51.34 & \cellcolor{LightBlue}38.94 & \cellcolor{LightBlue}30.32 & 46.19 & \cellcolor{LightBlue}46.79 & 35.88 & 40.99 & \cellcolor{LightBlue}\textbf{9.51} \\
\bottomrule
\end{tabular}
}
\caption{\label{tab:korean_data_num}
Ablation study on korean data number in Case 1, 2 and 3. $\dagger$  indicates that the model was trained using all 10 languages of CONLIPA, but with a different number of samples of Korean. The \colorbox{LightBlue}{skyblue boxes} indicate better performance compared to the baseline, and the \textbf{bold} text represents the best performance for each case and language.}.
\vspace{-1em}
\end{table*}

\section{Experimental Result with single language}
We present the quantitative result of all three cases in Table \ref{tab:single-table-all}. The method using phoneme representation outperforms in Case 1 and Case 2 in terms of average F1 score(\%) and demonstrates more stable results with a lower standard deviation.

\begin{table*}
\centering
\resizebox{\textwidth}{!}{
\begin{tabular}{c|cccccccccccccccc}
\toprule
Case & Model & \multicolumn{13}{c}{Languages} & AVG & STD \\
\midrule
\multirow{12}{*}{CASE 1}&&sin & som & mri & quy & uig & aii & kin & ilo & & & & & & & \\ \cmidrule{2-17}
 & XPhoneBERT & 43.61 & 38.91 & 38.07 & 51.90 & 44.82 & 31.03 & 49.67 & 73.05 & & & & & & 46.38 & 12.67 \\ \cmidrule{2-17}
&Swahili & \cellcolor{LightBlue}\textbf{44.74} & 38.71 & \cellcolor{LightBlue}38.12 & \cellcolor{LightBlue}53.66 & \cellcolor{LightBlue}44.89 & \cellcolor{LightBlue}31.65 & 49.43 & \cellcolor{LightBlue}73.24 & & & & & & \cellcolor{LightBlue}46.81 & 12.71 \\
&Indonesian & \cellcolor{LightBlue}44.43 & \cellcolor{LightBlue}39.05 & \cellcolor{LightBlue}\textbf{39.00} & \cellcolor{LightBlue}\textbf{55.53} & \cellcolor{LightBlue}44.84 & \cellcolor{LightBlue}32.54 & 49.43 & 72.46 & & & & & & \cellcolor{LightBlue}47.16 & \cellcolor{LightBlue}12.39 \\
&Hindi & \cellcolor{LightBlue}44.62 & 38.53 & \cellcolor{LightBlue}38.08 & \cellcolor{LightBlue}53.69 & \cellcolor{LightBlue}44.97 & 30.98 & 49.28 & \cellcolor{LightBlue}\textbf{73.25} & & & & & & \cellcolor{LightBlue}46.68 & 12.85 \\
&Mandarin & \cellcolor{LightBlue}44.37 & \cellcolor{LightBlue}\textbf{39.20} & \cellcolor{LightBlue}38.56 & \cellcolor{LightBlue}53.61 & \cellcolor{LightBlue}45.00 & \cellcolor{LightBlue}31.28 & 49.63 & 72.66 & & & & & & \cellcolor{LightBlue}46.79 & \cellcolor{LightBlue}12.53 \\
&Arabic & \cellcolor{LightBlue}44.46 & \cellcolor{LightBlue}39.11 & \cellcolor{LightBlue}38.55 & \cellcolor{LightBlue}55.02 & \cellcolor{LightBlue}44.90 & \cellcolor{LightBlue}32.56 & 49.4 & 72.71 & & & & & & \cellcolor{LightBlue}47.09 & \cellcolor{LightBlue}12.44 \\
&Vietnamese & \cellcolor{LightBlue}44.53 & \cellcolor{LightBlue}39.07 & 38.03 & \cellcolor{LightBlue}55.31 & \cellcolor{LightBlue}44.95 & \cellcolor{LightBlue}31.94 & \cellcolor{LightBlue}\textbf{50.10} & 72.69 & & & & & & \cellcolor{LightBlue}47.08 & \cellcolor{LightBlue}12.64 \\
&Thai & \cellcolor{LightBlue}44.61 & \cellcolor{LightBlue}39.15 & 37.94 & \cellcolor{LightBlue}54.53 & \cellcolor{LightBlue}\textbf{45.25} & \cellcolor{LightBlue}31.94 & \cellcolor{LightBlue}49.89 & 72.42 & & & & & & \cellcolor{LightBlue}46.97 & \cellcolor{LightBlue}12.48 \\
&Tamil & \cellcolor{LightBlue}44.43 & \cellcolor{LightBlue}39.07 & 37.95 & \cellcolor{LightBlue}54.68 & \cellcolor{LightBlue}45.00 & 30.75 & \cellcolor{LightBlue}50.01 & 72.81 & & & & & & \cellcolor{LightBlue}46.84 & \cellcolor{LightBlue}\textbf{12.01} \\
&Turkish & \cellcolor{LightBlue}44.62 & 38.89 & \cellcolor{LightBlue}38.22 & \cellcolor{LightBlue}54.93 & \cellcolor{LightBlue}44.98 & 30.81 & \cellcolor{LightBlue}50.09 & \cellcolor{LightBlue}73.24 & & & & & & \cellcolor{LightBlue}46.97 & 12.96 \\
&Korean & \cellcolor{LightBlue}44.57 & 38.51 & \cellcolor{LightBlue}38.75 & \cellcolor{LightBlue}55.48 & \cellcolor{LightBlue}44.86 & \cellcolor{LightBlue}\textbf{33.56} & 49.5 & 72.5 &&&&&& \cellcolor{LightBlue}\textbf{47.22} & \cellcolor{LightBlue}12.30  \\
\midrule
\multirow{12}{*}{CASE 2} && epo & khm & tuk & amh & mlt & ori & san & ina & grn & bel & kur & snd & &  & \\ \cmidrule{2-17}
 & XPhoneBERT & 75.26 & 31.86 & 61.17 & 44.85 & 52.58 & \textbf{40.73} & \textbf{59.42} & 68.68 & 49.95 & 77.61 & 52.95 & 47.28 & & 55.20 & 13.83  \\ \cmidrule{2-17}
&Swahili& 73.34 & \cellcolor{LightBlue}\textbf{38.81} & 58.95 & \cellcolor{LightBlue}51.75 & \cellcolor{LightBlue}55.16 & 40.54 & 54.62 & \cellcolor{LightBlue}70.00 & \cellcolor{LightBlue}54.23 & 72.33 & 51.76 & \cellcolor{LightBlue}47.71 && \cellcolor{LightBlue}55.77 & \cellcolor{LightBlue}11.35 \\
&Indonesian& 73.47 & \cellcolor{LightBlue}38.71 & 58.98 & \textbf{52.25} & \cellcolor{LightBlue}55.02 & 40.04 & 54.45 & \cellcolor{LightBlue}70.05 & \cellcolor{LightBlue}54.88 & 72.33 & 51.72 & \cellcolor{LightBlue}47.87 & & \cellcolor{LightBlue}55.81 & \cellcolor{LightBlue}11.42 \\
&Hindi& 73.3 & \cellcolor{LightBlue}38.62 & 58.79 & \cellcolor{LightBlue}51.61 & \cellcolor{LightBlue}55.06 & 40.62 & 54.79 & \cellcolor{LightBlue}69.88 & \cellcolor{LightBlue}54.35 & 72.32 & 52.04 & \cellcolor{LightBlue}47.71 & & \cellcolor{LightBlue}55.76 & \cellcolor{LightBlue}\textbf{11.33} \\
&Mandarin& 73.41 & \cellcolor{LightBlue}38.62 & 59.07 & \cellcolor{LightBlue}51.76 & \cellcolor{LightBlue}55.31 & 40.07 & 54.38 & \cellcolor{LightBlue}70.07 & \cellcolor{LightBlue}54.68 & 72.36 & 51.63 & \cellcolor{LightBlue}48.07 & & \cellcolor{LightBlue}55.79 & \cellcolor{LightBlue}11.43\\
&Arabic& 73.46 & \cellcolor{LightBlue}38.66 & 58.99 & \cellcolor{LightBlue}52.20 & \cellcolor{LightBlue}55.10 & 39.97 & 54.17 & \cellcolor{LightBlue}70.01 & \cellcolor{LightBlue}54.85 & 72.27 & 51.80 & \cellcolor{LightBlue}47.76 & & \cellcolor{LightBlue}55.77 & \cellcolor{LightBlue}11.43 \\
&Vietnamese& 73.54 & \cellcolor{LightBlue}38.56 & 59.23 & \cellcolor{LightBlue}51.93 & \cellcolor{LightBlue}\textbf{55.41} & 39.94 & 54.51 & \cellcolor{LightBlue}70.07 & \cellcolor{LightBlue}54.94 & 72.36 & 52.15 & \cellcolor{LightBlue}\textbf{48.15} & & \cellcolor{LightBlue}\textbf{55.90} & \cellcolor{LightBlue}11.45\\
&Thai& 73.48 & \cellcolor{LightBlue}38.68 & 59.25 & \cellcolor{LightBlue}51.84 & \cellcolor{LightBlue}\textbf{55.41} & 39.91 & 54.24 & \cellcolor{LightBlue}70.1 & \cellcolor{LightBlue}54.89 & 72.4 & 52.13 & \cellcolor{LightBlue}48.04 & & \cellcolor{LightBlue}55.86 & \cellcolor{LightBlue}11.45 \\
&Tamil& 73.41 & \cellcolor{LightBlue}38.64 & 59.19 & \cellcolor{LightBlue}51.73 & \cellcolor{LightBlue}55.28 & 40.12 & 54.38 & \cellcolor{LightBlue}70.08 & \cellcolor{LightBlue}54.88 & 72.29 & 52.08 & \cellcolor{LightBlue}47.94 & & \cellcolor{LightBlue}55.84 & \cellcolor{LightBlue}11.41 \\
&Turkish& 73.30 & \cellcolor{LightBlue}\textbf{38.81} & 59.32 & \cellcolor{LightBlue}51.63 & \cellcolor{LightBlue}55.13 & 40.24 & 55.06 & \cellcolor{LightBlue}70.01 & \cellcolor{LightBlue}\textbf{55.14} & 72.36 & 52.33 & \cellcolor{LightBlue}47.34 & & \cellcolor{LightBlue}55.89 & \cellcolor{LightBlue}11.39 \\
&Korean& 73.66 & \cellcolor{LightBlue}38.73 & 58.87 & \cellcolor{LightBlue}51.78 & \cellcolor{LightBlue}54.8 & 40.01 & 54.74 & \cellcolor{LightBlue}\textbf{70.18} & \cellcolor{LightBlue}55.04 & 72.21 & 51.63 & 47.15 && \cellcolor{LightBlue}55.73 & \cellcolor{LightBlue}11.51 \\
\midrule
\multirow{12}{*}{CASE 3}&&tgk & yor & mar & jav & urd & msa & ceb & hrv & mal & tel & uzb & pan & kir &  & \\ \cmidrule{2-17}
 & XPhoneBERT & 48.93 & 50.87 & 35.12 & 45.98 & 33.37 & 61.76 & 58.72 & 58.76 & 32.52 & 28.93 & 60.92 & 43.85 & 35.95 & 45.82 & 11.85  \\ \cmidrule{2-17}
&Swahili& 48.35 & \cellcolor{LightBlue}51.09 & \cellcolor{LightBlue}37.65 & \cellcolor{LightBlue}46.04 & \cellcolor{LightBlue}33.54 & 61.67 & 58.66 & 58.66 & \cellcolor{LightBlue}38.79 & \cellcolor{LightBlue}31.78 & \cellcolor{LightBlue}60.98 & \cellcolor{LightBlue}47.44 & \cellcolor{LightBlue}37.36 & \cellcolor{LightBlue}47.08 & \cellcolor{LightBlue}10.66\\
&Indonesian& \cellcolor{LightBlue}49.03 & 50.55 & \cellcolor{LightBlue}38.13 & \cellcolor{LightBlue}\textbf{46.22} & 33.22 & \cellcolor{LightBlue}62.01 & \cellcolor{LightBlue}59.04 & \cellcolor{LightBlue}58.84 & \cellcolor{LightBlue}39.27 & \cellcolor{LightBlue}32.22 & 60.83 & \cellcolor{LightBlue}\textbf{49.13} & \cellcolor{LightBlue}37.78 & \cellcolor{LightBlue}\textbf{47.41} & \cellcolor{LightBlue}10.62 \\
&Hindi& 48.56 & \cellcolor{LightBlue}51.35 & \cellcolor{LightBlue}37.65 & \cellcolor{LightBlue}46.03 & \cellcolor{LightBlue}\textbf{33.56} & 61.66 & 58.35 & 58.56 & \cellcolor{LightBlue}38.57 & \cellcolor{LightBlue}31.78 & \cellcolor{LightBlue}61.01 & \cellcolor{LightBlue}47.40 & \cellcolor{LightBlue}37.42 & \cellcolor{LightBlue}47.07 & \cellcolor{LightBlue}10.64 \\
&Mandarin& 48.92 & 50.78 & \cellcolor{LightBlue}38.11 & \cellcolor{LightBlue}46.09 & 33.21 & \cellcolor{LightBlue}62.00 & \cellcolor{LightBlue}58.98 & \cellcolor{LightBlue}58.95 & \cellcolor{LightBlue}\textbf{39.34} & \cellcolor{LightBlue}32.21 & \cellcolor{LightBlue}60.94 & \cellcolor{LightBlue}48.27 & \cellcolor{LightBlue}37.94 & \cellcolor{LightBlue}47.36 & \cellcolor{LightBlue}10.62 \\
&Arabic& \cellcolor{LightBlue}49.12 & 50.67 & \cellcolor{LightBlue}38.01 & \cellcolor{LightBlue}46.17 & 33.11 & \cellcolor{LightBlue}61.99 & \cellcolor{LightBlue}58.94 & \cellcolor{LightBlue}58.81 & \cellcolor{LightBlue}39.15 & \cellcolor{LightBlue}32.20 & \cellcolor{LightBlue}60.97 & \cellcolor{LightBlue}48.21 & \cellcolor{LightBlue}37.96 & \cellcolor{LightBlue}47.33 & \cellcolor{LightBlue}10.63 \\
&Vietnamese& \cellcolor{LightBlue}\textbf{49.20} & \cellcolor{LightBlue}50.90 & \cellcolor{LightBlue}\textbf{38.14} & \cellcolor{LightBlue}46.01 & 32.98 & \cellcolor{LightBlue}62.07 & \cellcolor{LightBlue}58.81 & \cellcolor{LightBlue}58.95 & \cellcolor{LightBlue}39.09 & \cellcolor{LightBlue}\textbf{32.45} & 60.92 & \cellcolor{LightBlue}48.54 & \cellcolor{LightBlue}\textbf{38.11} & \cellcolor{LightBlue}47.40 & \cellcolor{LightBlue}10.62 \\
&Thai& \cellcolor{LightBlue}49.05 & 50.72 & \cellcolor{LightBlue}38.08 & \cellcolor{LightBlue}46.02 & 33.04 & \cellcolor{LightBlue}62.11 & \cellcolor{LightBlue}\textbf{59.13} & \cellcolor{LightBlue}\textbf{58.98} & \cellcolor{LightBlue}39.17 & \cellcolor{LightBlue}32.35 & \cellcolor{LightBlue}61.03 & \cellcolor{LightBlue}48.47 & \cellcolor{LightBlue}37.99 & \cellcolor{LightBlue}47.40 & \cellcolor{LightBlue}10.67\\
&Tamil& \cellcolor{LightBlue}49.00 & 50.80 & \cellcolor{LightBlue}38.05 & \cellcolor{LightBlue}46.06 & 33.29 & \cellcolor{LightBlue}62.00 & 58.68 & \cellcolor{LightBlue}58.90 & \cellcolor{LightBlue}39.14 & \cellcolor{LightBlue}32.20 & \cellcolor{LightBlue}61.01 & \cellcolor{LightBlue}48.19 & \cellcolor{LightBlue}37.92 & \cellcolor{LightBlue}47.33 & \cellcolor{LightBlue}\textbf{10.61} \\
&Turkish& 48.26 & \cellcolor{LightBlue}51.14 & \cellcolor{LightBlue}37.69 & 45.97 & \cellcolor{LightBlue}33.54 & 61.73 & 58.37 & 58.71 & \cellcolor{LightBlue}38.55 & \cellcolor{LightBlue}31.85 & \cellcolor{LightBlue}\textbf{61.16} & \cellcolor{LightBlue}47.41 & \cellcolor{LightBlue}37.38 & \cellcolor{LightBlue}47.06 & \cellcolor{LightBlue}10.67 \\
&Korean& 48.41 & \cellcolor{LightBlue}\textbf{51.48} & \cellcolor{LightBlue}37.76 & 45.57 & 32.59 & \cellcolor{LightBlue}\textbf{62.25} & 57.4 & 58.71 & \cellcolor{LightBlue}38.35 & \cellcolor{LightBlue}31.91 & 60.54 & \cellcolor{LightBlue}48.04 & \cellcolor{LightBlue}37.66 & \cellcolor{LightBlue}46.97 & \cellcolor{LightBlue}10.68 \\
\bottomrule
\end{tabular}
}
\caption{\label{tab:single-table-all}
Ablation study on each language in case 1,2,3.
\vspace{-1em}
}
\end{table*}

\section{Ablation on the Temperature Coefficient}
As discussed in paper \cite{kim2025temperature}, InfoNCE loss~\cite{DBLP:journals/corr/abs-1807-03748, chen2020simple} is commonly employed in contrastive learning since it facilitates learning data representations by capturing the similarities between pairs. While InfoNCE loss plays a crucial role~\cite{wang2021understanding, zhangdoes}, it requires the tuning of a temperature parameter. This critical hyperparameter modifies the similarity scores and governs the intensity of penalties applied to difficult negative samples\cite{wang2021understanding}. This temperature coefficient is represented by $\tau$ in equation \ref{eqn:InfoNCE_loss} of Section \ref{section_4}.

To identify an optimal temperature, we conducted an ablation study by varying only the temperature coefficient. The study was performed using 512 Korean data samples, along with data from 10 other languages. As shown in Figure \ref{tab:temperature_table_all}, a temperature coefficient of 0.1 yielded the best performance in our experiment.

\begin{table*}
\centering
\resizebox{\textwidth}{!}{
\begin{tabular}{c|cccccccccccccccc}
\toprule
Case & Temperature & \multicolumn{13}{c}{Languages} & AVG & STD \\
\midrule
\multirow{12}{*}{CASE 1}&&sin & som & mri & quy & uig & aii & kin & ilo & & & & & & & \\ \cmidrule{2-17}
&0.01 & 37.99 & \cellcolor{LightBlue}\textbf{42.86} & \cellcolor{LightBlue}39.96 & 49.51 & \cellcolor{LightBlue}\textbf{49.15} & 27.4 & \cellcolor{LightBlue}\textbf{53.71} & 72.49 &&&&&& \cellcolor{LightBlue}46.63 & 13.29  \\
&0.05 & \cellcolor{LightBlue}46.05 & \cellcolor{LightBlue}39.14 & \cellcolor{LightBlue}39.27 & \cellcolor{LightBlue}54.87 & \cellcolor{LightBlue}44.96 & \cellcolor{LightBlue}32.59 & 49.38 & 72.02 &&&&&& \cellcolor{LightBlue}47.29 & \cellcolor{LightBlue}12.11  \\
&0.1 & \cellcolor{LightBlue}45.69 & 38.7 & \cellcolor{LightBlue}39.67 & \cellcolor{LightBlue}\textbf{57.70} & \cellcolor{LightBlue}45.17 & \cellcolor{LightBlue}34.92 & \cellcolor{LightBlue}50.58 & \cellcolor{LightBlue}\textbf{73.35} &&&&&& \cellcolor{LightBlue}\textbf{48.22} & \cellcolor{LightBlue}12.44  \\
&0.15 & \cellcolor{LightBlue}\textbf{46.62} & 36.92 & \cellcolor{LightBlue}40.10 & \cellcolor{LightBlue}54.49 & 44.17 & \cellcolor{LightBlue}37.74 & 48.50 & 72.74 &&&&&& \cellcolor{LightBlue}47.66 & \cellcolor{LightBlue}11.71 \\
&0.2 & \cellcolor{LightBlue}46.09 & 36.25 & \cellcolor{LightBlue}\textbf{40.15} & \cellcolor{LightBlue}52.45 & 42.44 & \cellcolor{LightBlue}38.01 & 47.84 & \cellcolor{LightBlue}73.33 &&&&&& \cellcolor{LightBlue}47.07 & \cellcolor{LightBlue}11.88  \\
&0.3 & \cellcolor{LightBlue}45.73 & 36.04 & \cellcolor{LightBlue}40.05 & 51.32 & 41.57 & \cellcolor{LightBlue}38.83 & 47.28 & 72.97 &&&&&& \cellcolor{LightBlue}46.72 & \cellcolor{LightBlue}11.70 \\
&0.4 & \cellcolor{LightBlue}45.62 & 35.47 & \cellcolor{LightBlue}38.73 & 48.45 & 41.08 & \cellcolor{LightBlue}39.76 & 47.8 & 72.81 &&&&&& 46.22 & \cellcolor{LightBlue}11.68  \\
&0.5 & \cellcolor{LightBlue}45.71 & 35.38 & \cellcolor{LightBlue}38.67 & 48.8 & 40.82 & \cellcolor{LightBlue}39.17 & 47.8 & 72.74 &&&&&& 46.14 & \cellcolor{LightBlue}11.75  \\
&0.6 & \cellcolor{LightBlue}45.82 & 35.45 & \cellcolor{LightBlue}39.69 & 49.35 & 41.14 & \cellcolor{LightBlue}39.90 & 47.56 & 72.71 &&&&&& \cellcolor{LightBlue}46.45 & \cellcolor{LightBlue}11.57  \\
&0.7 & \cellcolor{LightBlue}45.71 & 35.41 & \cellcolor{LightBlue}38.51 & 48.68 & 41.28 & \cellcolor{LightBlue}39.98 & 47.81 & 72.20 &&&&&& 46.20 & \cellcolor{LightBlue}\textbf{11.49} \\
&0.8 & \cellcolor{LightBlue}45.76 & 35.27 & \cellcolor{LightBlue}38.52 & 48.67 & 40.41 & \cellcolor{LightBlue}39.17 & 48.11 & 72.07 &&&&&& 46.00 & \cellcolor{LightBlue}11.59 \\
&0.9 & \cellcolor{LightBlue}45.35 & 35.34 & \cellcolor{LightBlue}39.50 & 49.22 & 41.07 & \cellcolor{LightBlue}39.93 & 47.56 & 72.63 &&&&&& 46.33 & \cellcolor{LightBlue}11.58  \\
&1.0 & \cellcolor{LightBlue}46.07 & 35.24 & \cellcolor{LightBlue}38.37 & 48.34 & 41.16 & \cellcolor{LightBlue}\textbf{39.99} & 47.77 & 72.12 &&&&&& 46.13 & \cellcolor{LightBlue}11.49 \\
\midrule
\multirow{12}{*}{CASE 2} && epo & khm & tuk & amh & mlt & ori & san & ina & grn & bel & kur & snd & &  & \\ \cmidrule{2-17}
&0.01 & 72.95 & \cellcolor{LightBlue}34.90 & 59.68 & \cellcolor{LightBlue}51.80 & \cellcolor{LightBlue}\textbf{57.01} & 35.83 & \textbf{55.04} & \cellcolor{LightBlue}\textbf{71.36} & \cellcolor{LightBlue}53.09 & \textbf{72.91} & 46.56 & \cellcolor{LightBlue}48.52 && 54.97 & \cellcolor{LightBlue}12.92 \\
&0.05 &73.70 & \cellcolor{LightBlue}38.54 & 59.87 & \cellcolor{LightBlue}51.11 & \cellcolor{LightBlue}54.43 & 39.20 & 55.01 & \cellcolor{LightBlue}71.18 & \cellcolor{LightBlue}\textbf{53.75} & 72.71 & 51.58 & \cellcolor{LightBlue}49.31 && \cellcolor{LightBlue}55.87 & \cellcolor{LightBlue}11.76 \\
&0.1 & \textbf{74.11} & \cellcolor{LightBlue}39.95 & \textbf{60.97} & \cellcolor{LightBlue}50.14 & \cellcolor{LightBlue}54.03 & 40.10 & 53.49 & \cellcolor{LightBlue}70.73 & \cellcolor{LightBlue}53.17 & 72.72 & 52.00 & \cellcolor{LightBlue}48.44 && \cellcolor{LightBlue}55.82 & \cellcolor{LightBlue}11.62 \\
&0.15 & 74.05 & \cellcolor{LightBlue}42.78 & 60.45 & \cellcolor{LightBlue}51.14 & 52.32 & \cellcolor{LightBlue}41.34 & 52.77 & \cellcolor{LightBlue}71.22 & \cellcolor{LightBlue}52.12 & 72.54 & 51.95 & \cellcolor{LightBlue}49.20 && \cellcolor{LightBlue}55.99 & \cellcolor{LightBlue}11.14 \\
&0.2 & 74.08 & \cellcolor{LightBlue}44.28 & 60.38 & \cellcolor{LightBlue}50.84 & 51.91 & \cellcolor{LightBlue}42.05 & 52.57 & \cellcolor{LightBlue}70.80 & \cellcolor{LightBlue}51.58 & 72.61 & 51.56 & \cellcolor{LightBlue}49.00 && \cellcolor{LightBlue}55.97 & \cellcolor{LightBlue}10.94 \\
&0.3 & 73.98 & \cellcolor{LightBlue}45.82 & 60.26 & \cellcolor{LightBlue}51.41 & 51.24 & \cellcolor{LightBlue}42.98 & 52.55 & \cellcolor{LightBlue}70.39 & \cellcolor{LightBlue}51.14 & 72.44 & 51.98 & \cellcolor{LightBlue}48.78 && \cellcolor{LightBlue}56.08 & \cellcolor{LightBlue}10.61 \\
&0.4 & 73.79 & \cellcolor{LightBlue}45.82 & 59.18 & \cellcolor{LightBlue}52.46 & 51.59 & \cellcolor{LightBlue}43.21 & 52.29 & \cellcolor{LightBlue}70.77 & \cellcolor{LightBlue}52.14 & 72.37 & \cellcolor{LightBlue}52.98 & \cellcolor{LightBlue}\textbf{49.57} && \cellcolor{LightBlue}\textbf{56.35} & \cellcolor{LightBlue}10.40 \\
&0.5 & 73.78 & \cellcolor{LightBlue}45.84 & 59.15 & \cellcolor{LightBlue}52.45 & 51.51 & \cellcolor{LightBlue}43.21 & 52.28 & \cellcolor{LightBlue}70.63 & \cellcolor{LightBlue}51.82 & 72.31 & 53.01 & \cellcolor{LightBlue}49.26 && \cellcolor{LightBlue}56.27 & \cellcolor{LightBlue}10.40 \\
&0.6 & 73.86 & \cellcolor{LightBlue}46.17 & 60.01 & \cellcolor{LightBlue}51.81 & 50.93 & \cellcolor{LightBlue}43.17 & 51.76 & \cellcolor{LightBlue}70.30 & \cellcolor{LightBlue}50.98 & 72.30 & 52.02 & \cellcolor{LightBlue}49.09 && \cellcolor{LightBlue}56.03 & \cellcolor{LightBlue}10.52 \\
&0.7 & 73.85 & \cellcolor{LightBlue}46.33 & 60.46 & \cellcolor{LightBlue}52.40 & 51.66 & \cellcolor{LightBlue}42.84 & 52.34 & \cellcolor{LightBlue}70.78 & \cellcolor{LightBlue}50.96 & 72.36 & 52.12 & \cellcolor{LightBlue}48.95 && \cellcolor{LightBlue}56.25 & \cellcolor{LightBlue}10.56 \\
&0.8 & 73.73 & \cellcolor{LightBlue}45.61 & 59.16 & \cellcolor{LightBlue}52.32 & 51.58 & \cellcolor{LightBlue}\textbf{43.21} & 52.24 & \cellcolor{LightBlue}70.28 & \cellcolor{LightBlue}51.78 & 72.32 & \cellcolor{LightBlue}\textbf{53.08} & \cellcolor{LightBlue}49.18 && \cellcolor{LightBlue}56.21 & \cellcolor{LightBlue}\textbf{10.38} \\
&0.9 & 73.81 & \cellcolor{LightBlue}45.91 & 59.98 & \cellcolor{LightBlue}51.84 & 50.90 & \cellcolor{LightBlue}43.04 & 51.80 & \cellcolor{LightBlue}70.27 & \cellcolor{LightBlue}51.04 & 72.32 & 52.15 & \cellcolor{LightBlue}48.71 && \cellcolor{LightBlue}55.98 & \cellcolor{LightBlue}10.56 \\
&1.0 & 73.83 & \cellcolor{LightBlue}\textbf{46.84} & 59.63 & \cellcolor{LightBlue}\textbf{52.46} & 51.61 & \cellcolor{LightBlue}42.91 & 52.56 & \cellcolor{LightBlue}70.76 & \cellcolor{LightBlue}51.31 & 72.31 & 52.18 & \cellcolor{LightBlue}48.93 && \cellcolor{LightBlue}56.28 & \cellcolor{LightBlue}10.44 \\
\midrule
\multirow{12}{*}{CASE 3}&&tgk & yor & mar & jav & urd & msa & ceb & hrv & mal & tel & uzb & pan & kir &  & \\ \cmidrule{2-17}
&0.01 & \cellcolor{LightBlue}49.89 & \textbf{50.81} & \cellcolor{LightBlue}38.58 & \cellcolor{LightBlue}46.08 & \cellcolor{LightBlue}\textbf{33.89} & 61.09 & \cellcolor{LightBlue}58.91 & \cellcolor{LightBlue}\textbf{61.24} & \cellcolor{LightBlue}39.41 & \cellcolor{LightBlue}\textbf{33.90} & 60.15 & \cellcolor{LightBlue}48.14 & \cellcolor{LightBlue}38.13 & \cellcolor{LightBlue}47.71 & \cellcolor{LightBlue}\textbf{10.35} \\
&0.05 & \cellcolor{LightBlue}\textbf{50.14} & 49.08 & \cellcolor{LightBlue}38.90 & 45.63 & 32.24 & \cellcolor{LightBlue}62.13 & \cellcolor{LightBlue}\textbf{59.37} & \cellcolor{LightBlue}59.74 & \cellcolor{LightBlue}\textbf{40.11} & \cellcolor{LightBlue}33.13 & \textbf{60.58} & \cellcolor{LightBlue}49.81 & \cellcolor{LightBlue}\textbf{39.62} & \cellcolor{LightBlue}\textbf{47.73} & \cellcolor{LightBlue}10.50 \\
&0.1 & 48.19 & 50.05 & \cellcolor{LightBlue}\textbf{38.97} & \cellcolor{LightBlue}\textbf{46.24} & 31.35 & \cellcolor{LightBlue}\textbf{62.83} & 58.16 & \cellcolor{LightBlue}59.17 & \cellcolor{LightBlue}39.50 & \cellcolor{LightBlue}32.57 & 60.38 & \cellcolor{LightBlue}49.48 & \cellcolor{LightBlue}39.37 & \cellcolor{LightBlue}47.40 & \cellcolor{LightBlue}10.61 \\
&0.15 & 47.19 & 47.85 & \cellcolor{LightBlue}38.41 & \cellcolor{LightBlue}46.08 & 30.27 & \cellcolor{LightBlue}62.52 & 58.43 & 58.57 & \cellcolor{LightBlue}39.45 & \cellcolor{LightBlue}32.86 & 59.42 & \cellcolor{LightBlue}50.17 & \cellcolor{LightBlue}38.00 & \cellcolor{LightBlue}46.86 & \cellcolor{LightBlue}10.66 \\
&0.2 & 45.62 & 47.96 & \cellcolor{LightBlue}38.03 & \cellcolor{LightBlue}46.19 & 30.11 & \cellcolor{LightBlue}62.44 & 58.24 & 57.91 & \cellcolor{LightBlue}38.95 & \cellcolor{LightBlue}32.25 & 59.41 & \cellcolor{LightBlue}50.98 & \cellcolor{LightBlue}37.35 & \cellcolor{LightBlue}46.57 & \cellcolor{LightBlue}10.79 \\
&0.3 & 44.99 & 47.35 & \cellcolor{LightBlue}38.10 & \cellcolor{LightBlue}46.23 & 29.87 & \cellcolor{LightBlue}62.09 & 58.13 & 57.70 & \cellcolor{LightBlue}38.99 & \cellcolor{LightBlue}32.35 & 58.96 & \cellcolor{LightBlue}50.59 & \cellcolor{LightBlue}36.90 & \cellcolor{LightBlue}46.33 & \cellcolor{LightBlue}10.70 \\
&0.4 & 45.38 & 46.90 & \cellcolor{LightBlue}38.22 & 45.15 & 30.32 & 61.33 & 57.82 & 57.88 & \cellcolor{LightBlue}39.69 & \cellcolor{LightBlue}32.75 & 58.89 & \cellcolor{LightBlue}51.01 & \cellcolor{LightBlue}37.09 & \cellcolor{LightBlue}46.34 & \cellcolor{LightBlue}10.44 \\
&0.5 & 44.94 & 47.15 & \cellcolor{LightBlue}38.13 & 45.10 & 30.29 & 61.31 & 57.76 & 57.82 & \cellcolor{LightBlue}39.62 & \cellcolor{LightBlue}32.73 & 58.84 & \cellcolor{LightBlue}50.93 & \cellcolor{LightBlue}36.09 & \cellcolor{LightBlue}46.21 & \cellcolor{LightBlue}10.52 \\
&0.6 & 45.24 & 47.21 & \cellcolor{LightBlue}38.05 & \cellcolor{LightBlue}46.23 & 29.79 & 61.67 & 57.92 & 57.54 & \cellcolor{LightBlue}39.09 & \cellcolor{LightBlue}32.37 & 58.51 & \cellcolor{LightBlue}51.19 & \cellcolor{LightBlue}36.67 & \cellcolor{LightBlue}46.27 & \cellcolor{LightBlue}10.61 \\
&0.7 & 45.41 & 46.80 & \cellcolor{LightBlue}38.31 & \cellcolor{LightBlue}46.05 & 29.76 & 61.60 & 58.35 & 57.79 & \cellcolor{LightBlue}39.60 & \cellcolor{LightBlue}32.82 & 58.36 & \cellcolor{LightBlue}51.96 & \cellcolor{LightBlue}36.86 & \cellcolor{LightBlue}46.44 & \cellcolor{LightBlue}10.58 \\
&0.8 & 44.00 & 46.94 & \cellcolor{LightBlue}38.15 & 45.27 & 30.08 & 61.21 & 57.72 & 57.72 & \cellcolor{LightBlue}39.80 & \cellcolor{LightBlue}32.81 & 58.64 & \cellcolor{LightBlue}51.39 & \cellcolor{LightBlue}35.96 & \cellcolor{LightBlue}46.13 & \cellcolor{LightBlue}10.52 \\
&0.9 & 44.95 & 47.00 & \cellcolor{LightBlue}38.08 & \cellcolor{LightBlue}46.10 & 29.78 & 61.50 & 58.00 & 57.50 & \cellcolor{LightBlue}39.20 & \cellcolor{LightBlue}32.34 & 58.48 & \cellcolor{LightBlue}51.36 & \cellcolor{LightBlue}36.09 & \cellcolor{LightBlue}46.18 & \cellcolor{LightBlue}10.64 \\
&1.0 & 45.33 & 46.83 & \cellcolor{LightBlue}38.30 & \cellcolor{LightBlue}46.04 & 29.68 & 61.56 & 58.31 & 57.75 & \cellcolor{LightBlue}39.69 & \cellcolor{LightBlue}32.83 & 58.25 & \cellcolor{LightBlue}\textbf{52.05} & \cellcolor{LightBlue}36.73 & \cellcolor{LightBlue}46.41 & \cellcolor{LightBlue}10.58 \\
\bottomrule
\end{tabular}
}
\caption{\label{tab:temperature_table_all}
Ablation study on contrastive learning temperature in case 1,2,3.
\vspace{-1em}
}
\end{table*}

\section{Visualization of Phonemic Representation}
\label{t_sne_visualization}
We analyzed the distance between eng-ori and eng-khm word pairs in Section \ref{cosine_similarity} of the main paper. Here, we also visualize the distribution of representations in a zero-shot setting, where phoneme input from a low-resource language is presented solely during inference, without prior exposure during training.

\begin{figure*}[t!]
    \centering
    \includegraphics[width=0.8\linewidth]{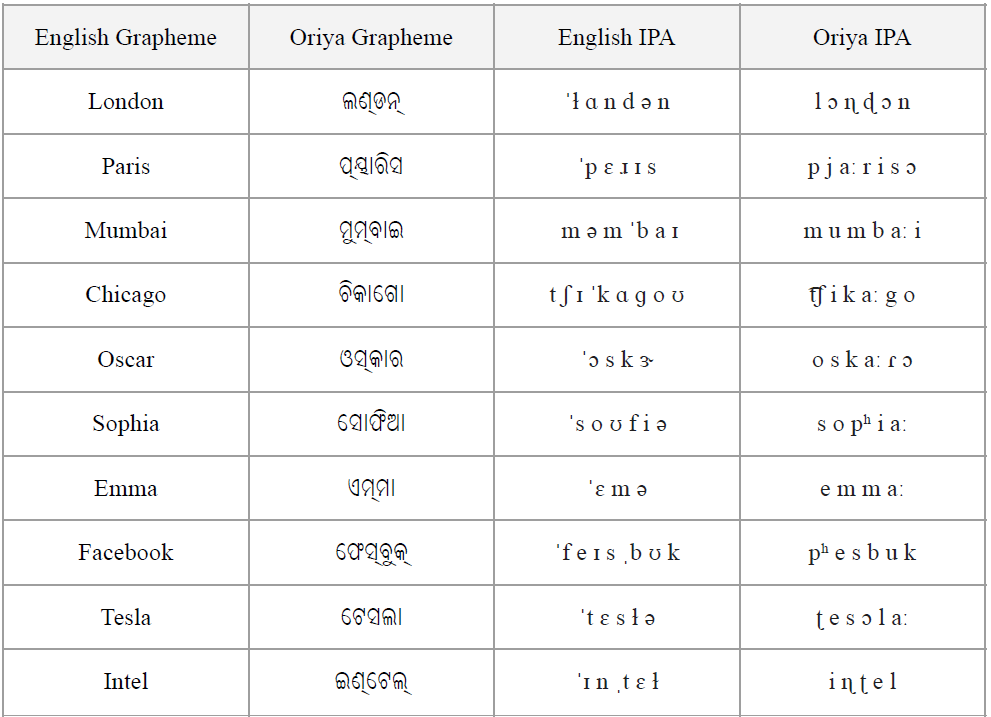}
    \caption{Ten eng-ori pairs with similar pronunciations}
    \label{fig:eng_oriya_pair}
\end{figure*}

\begin{figure*}[t!]
    \centering
    \includegraphics[width=0.8\linewidth]{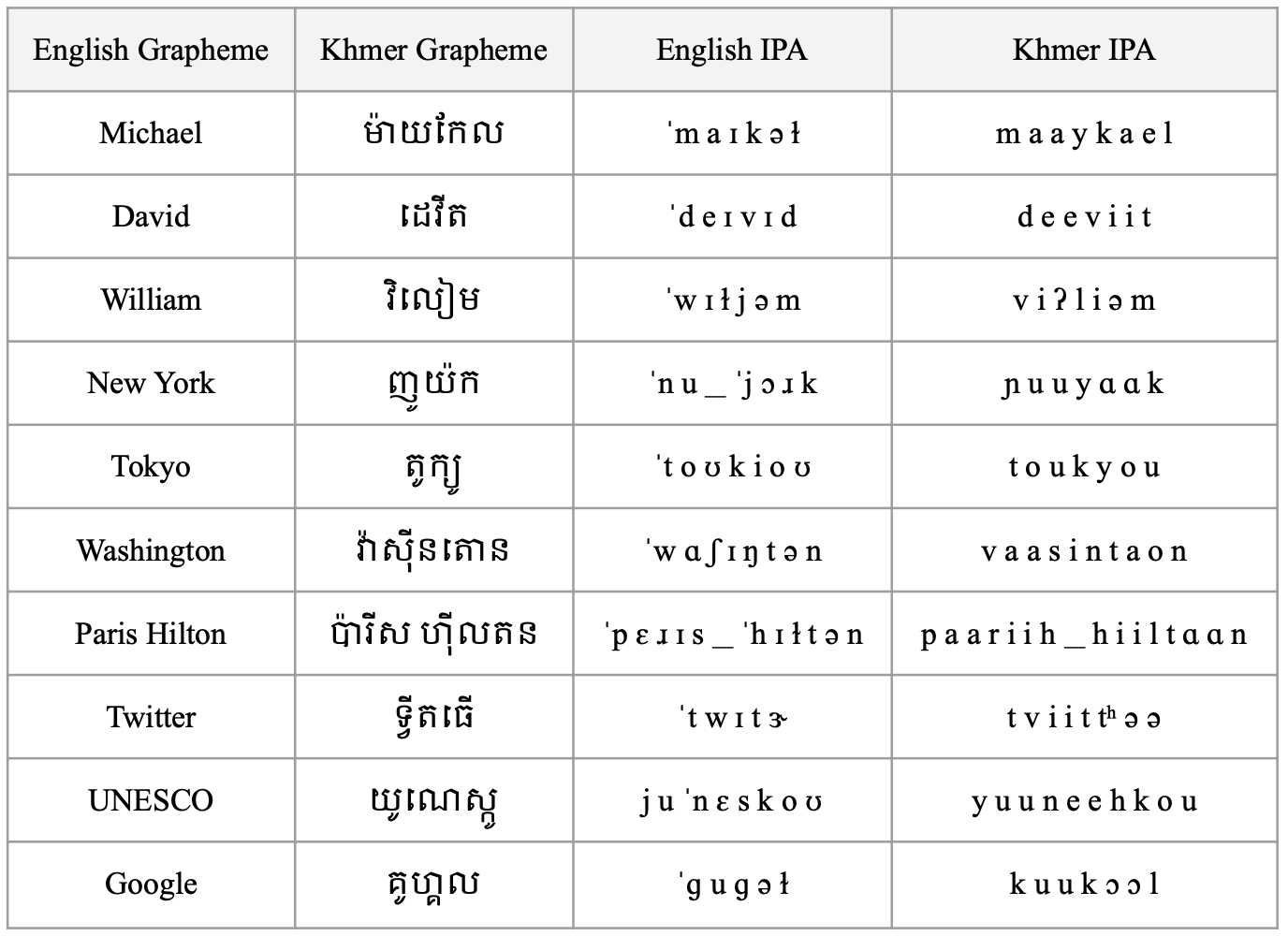}
    \caption{Ten eng-khm pairs with similar pronunciations}
    \label{fig:eng_khmer_pair}
\end{figure*}

We employed t-SNE to compare how the distribution of representations changes before and after IPA contrastive learning. For this study, we selected Oriya and Khmer as low-resource languages. We used IPA inputs corresponding to 10 English-target language pairs for both Oriya and Khmer, focusing on words with similar pronunciations. The selected words consisted of person, organization, and location named entities. Figures \ref{fig:eng_oriya_pair} and \ref{fig:eng_khmer_pair} present the 10 samples for Oriya and Khmer, respectively.

\begin{figure*}[h!]
    \centering
    \subfloat[eng-ori before training]{\includegraphics[width=0.48\linewidth]{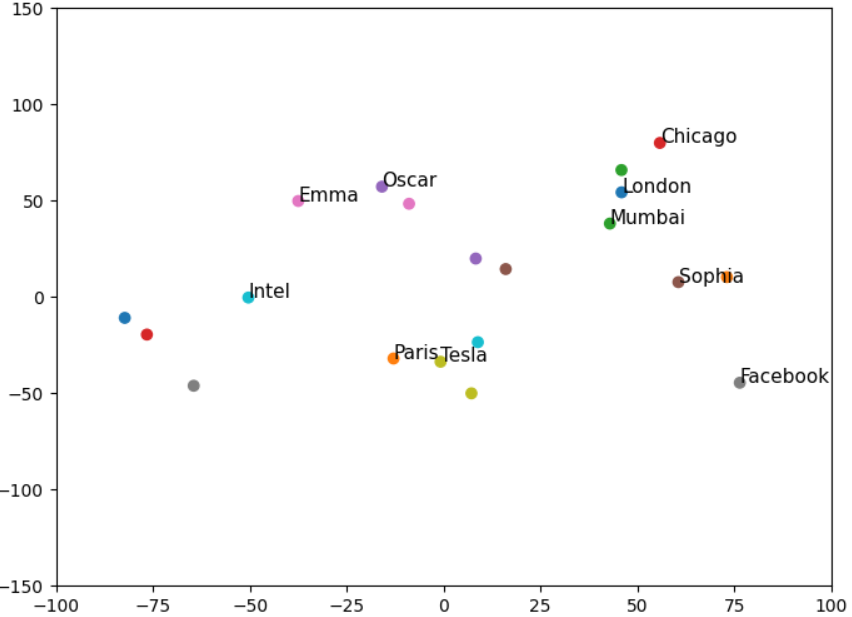}}
    \subfloat[eng-ori after training]{\includegraphics[width=0.48\linewidth]{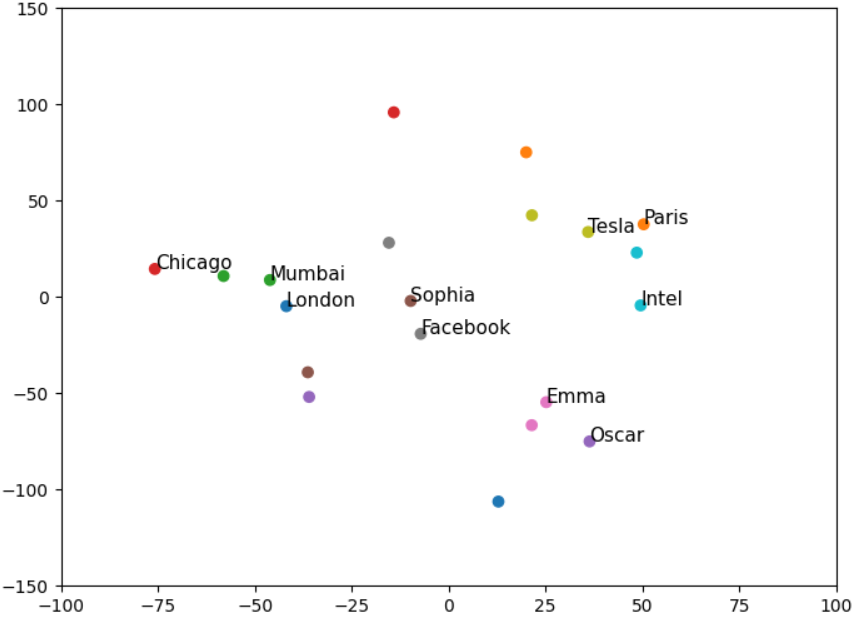}}
    \hfill
    \subfloat[eng-khm before training]
    {\includegraphics[width=0.48\linewidth]{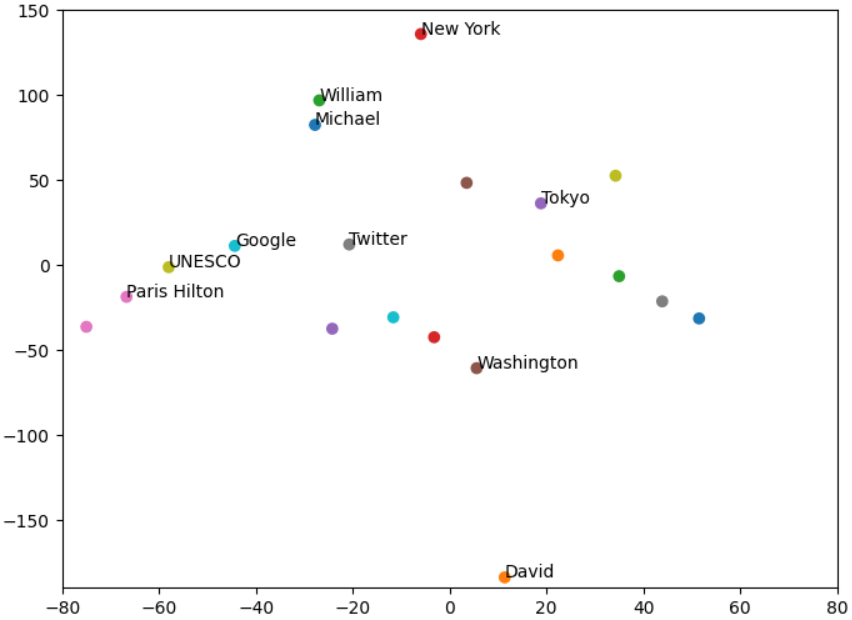}}
    \subfloat[eng-khm after training]{\includegraphics[width=0.48\linewidth]{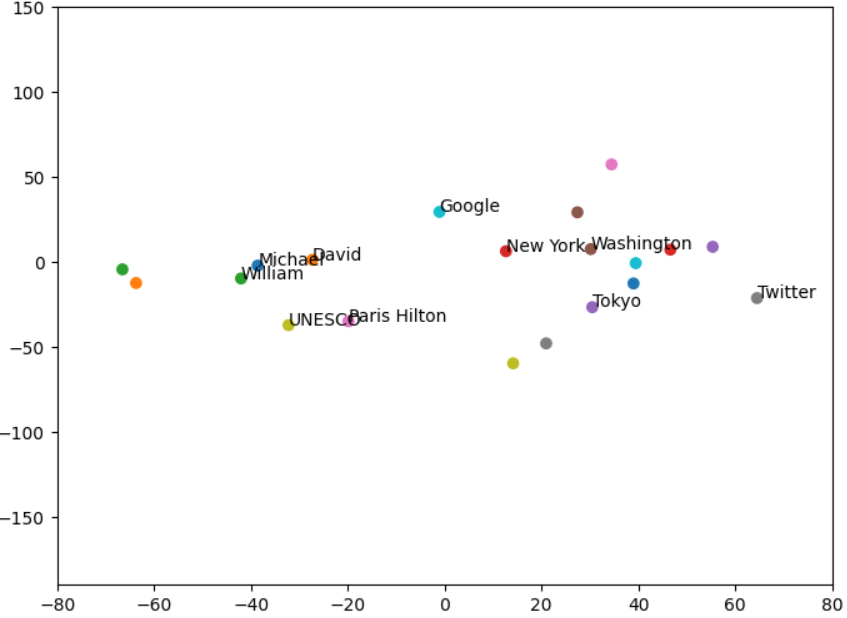}}
    \caption{\label{fig:tsne_oriya_khmer} t-SNE (perplexity=2) visualization using 10 eng-ori pairs and 10 eng-khm pairs. Panels (a) and (c) represent the results before IPA contrastive learning, while panels (b) and (d) show the results after learning. Dots of the same color indicate pairs of english and target language words with the same meaning.}
\end{figure*}

As shown in Figure \ref{fig:tsne_oriya_khmer}, we compared the t-SNE results before and after IPA contrastive learning using our CONLIPA dataset. The results before learning are shown in (a) and (c), while those after learning are shown in (b) and (d). In the figure, dots of the same color represent pairs of English and target language words with the same meaning, with only the English labels displayed.

In (a) and (c), most of the paired points were distant from each other. Since Oriya and Khmer are low-resource languages, even when input is given in IPA notation, there was a noticeable distance between the paired points. However, in (b) and (d), the distance between these points was significantly reduced.

Note that only the IPA representations of both English and the target language, rather than their grapheme notations, are used for visualization in this process. Additionally, it should be noted that the examples of Oriya and Khmer were not used for any training, such as pre-training with WikiANN or IPA contrastive learning, nor for zero-shot inference. These samples were created and used solely for cosine similarity calculation and t-SNE visualization purposes.

We configured the t-SNE with perplexity=2 and n\_iter=300 to generate the visualizations. To ensure a fair comparison, we standardized the axis ranges: for eng-ori, the x-axis range was [-100, 100] and the y-axis range was [-150, 150], while for eng-khm, the x-axis range was [-80, 80] and the y-axis range was [-190, 150].

